\documentclass[%
 reprint,
preprintnumbers,
nofootinbib,
nobibnotes,
 aps,
]{revtex4-2}

\usepackage[utf8]{inputenc}

\usepackage{xr-hyper}
\usepackage{url}

\usepackage{enumitem}
\usepackage{setspace}
\usepackage{amscd,amsmath,amssymb,amsfonts,amsthm,bbm,bm,latexsym,mathrsfs,mathtools}
\usepackage[subnum]{cases}	% multi-case equations with separate number for each
\usepackage{hyperref} 	
\hypersetup{unicode=true, colorlinks=false, urlcolor=blue, linkcolor=blue, filecolor=blue, citecolor=blue}
\usepackage{xcolor,subfigure,epsfig,rotating,graphics,graphicx} 
\graphicspath{{./figures/}}
\usepackage{float,array,multirow}
\usepackage{threeparttable}
\usepackage[width=0.92\linewidth, font={footnotesize}]{caption}
\usepackage{graphicx}
\newlength\replength
\newcommand\repfrac{.33}

\newcommand\rulewidth{.6pt}
\newcommand\tdashfill[1][\repfrac]{\cleaders\hbox to \replength{%
  \smash{\rule[\arraystretch\ht\strutbox]{\repfrac\replength}{\rulewidth}}}\hfill}

\newcommand\tdotfill[1][\repfrac]{\cleaders\hbox to \replength{%
  \smash{\raisebox{\arraystretch\dimexpr\ht\strutbox-.1ex\relax}{.}}}\hfill}

\makeatletter
\newcommand*{\addFileDependency}[1]{% argument=file name and extension
  \typeout{(#1)}
  \@addtofilelist{#1}
  \IfFileExists{#1}{}{\typeout{No file #1.}}
}
\makeatother

\newcommand*{\myexternaldocument}[1]{%
    \externaldocument{#1}%
    \addFileDependency{#1.tex}%
    \addFileDependency{#1.aux}%
}
\myexternaldocument{supplement_V1}
\begin{document}

% A roadmap for analysing arguments and debates

%\title{Online Arguments \& Debates: The Causal Rhetorics of a No-Deal Brexit}
\title{The Unfolding Structure of Arguments in Online Debates:\vspace{0.5em}\hspace{15em} The case of a No-Deal Brexit}
%The case of No-Deal Brexit Causal Rhetoric
%acronyms aond abbreviations

\newcommand{\method}{ROAD}
%\thanks{A footnote to the article title}

\author{Carlo R. M. A. Santagiustina}
 \email{carlo.santagiustina@unive.it}
\altaffiliation[Also at ]{Venice International University, Isola di San Servolo, 30133 Venice, Italy\\ 
\& Department of Economics, Ca' Foscari University of Venice, Cannaregio 873, 30123 Venice, Italy}%Lines break automatically or can be forced with \\
\author{Massimo Warglien}
 \email{warglien@unive.it}
\affiliation{Department of Management, Ca' Foscari University of Venice, Cannaregio 873, 30123 Venice, Italy}%

\date{\today}
%\collaboration{Collaboration}%\noaffiliation

%\keywords{Keyword1, Keyword2, Keyword3}

\begin{abstract}
%Example Abstract. Abstract must not include subheadings or citations. 
In the last decade, political debates have progressively shifted to social media. Rhetorical devices employed by online actors and factions that operate in these debating arenas can be captured and analysed to conduct a statistical reading of societal controversies and their argumentation dynamics. In this paper, we propose a five-step methodology, to extract, categorize and explore the latent argumentation structures of online debates. Using Twitter data about a ``no-deal'' Brexit, we focus on the expected effects in case of materialisation of this event.  First, we extract cause-effect claims contained in tweets using regular expressions that exploit verbs related to \textit{Creation}, \textit{Destruction} and \textit{Causation}. Second, we categorise extracted ``no-deal'' effects using a Structural Topic Model estimated on unigrams and bigrams. Third, we select controversial effect topics and explore within-topic argumentation differences between self-declared partisan user factions, i.e., \textit{Brexiteers} and \textit{Remainers}.  
We hence type topics using estimated covariate effects on topic propensities, then, using the topics correlation network, we study the aggregate topological structure of the debate to identify coherent topical constellations. 
Finally, we analyse the debate time dynamics and infer lead/follow relations among factions. Results show that the proposed methodology can be employed to perform a statistical rhetorics analysis of debates, and map the architecture of controversies across time. In particular, the ``no-deal'' Brexit debate on Twitter is shown to have a multifaceted assortative argumentation structure heavily characterized by factional constellations of arguments, as well as by polarized narrative frames invoked through verbs related to \textit{Creation} and \textit{Destruction}.
Our findings highlight the benefits of implementing a systemic approach to the analysis of debates, which allows the unveiling of topical and factional dependencies between causal arguments and rhetoric devices employed in online debates.
\end{abstract}

\keywords{online debates, topic modeling, Brexit, No-Deal}%Use showkeys class option if keyword

\maketitle

\section{\label{sec:intro}{Introduction}}

Online debates have become a major component of contemporary democratic life, involving millions of people in the expression of opinions on a vast range of topics\cite{gottfried2017changing,lazer2009life}. 
Debates are strictly associated with argumentation to speech acts that express participants’ opinions and try to affect other participants’ views by offering reasons, triggering frames, eliciting emotions\cite{wiesner2017debates}.  Debates can be finalized to deliberation, as in public assemblies, or more loosely directed to communicate and shape opinions on controversial subjects or issues. Online debates are typically of this second type. 
Online Social Media (OSM) platforms, like Twitter, Facebook and Reddit, are arenas where these lively debates nowadays take place. These virtual spaces are incessantly used by wide communities of users to gather information, communicate their thoughts and views concerning (realized or possible) events occurring at the national or global scale. Partisan and non-partisan participants to these online debates often state publicly their opinions about these events, their likelihood and expected effects. Therefore, online debates offer a privileged window on the expressed arguments of a large sample of the politically and socially active population. 

\smallskip\par
 \begin{figure*}
\centering
    \begin{tabular}{c}
\includegraphics[trim= 0mm 0mm 0mm 0mm,clip, width= 0.8\textwidth]{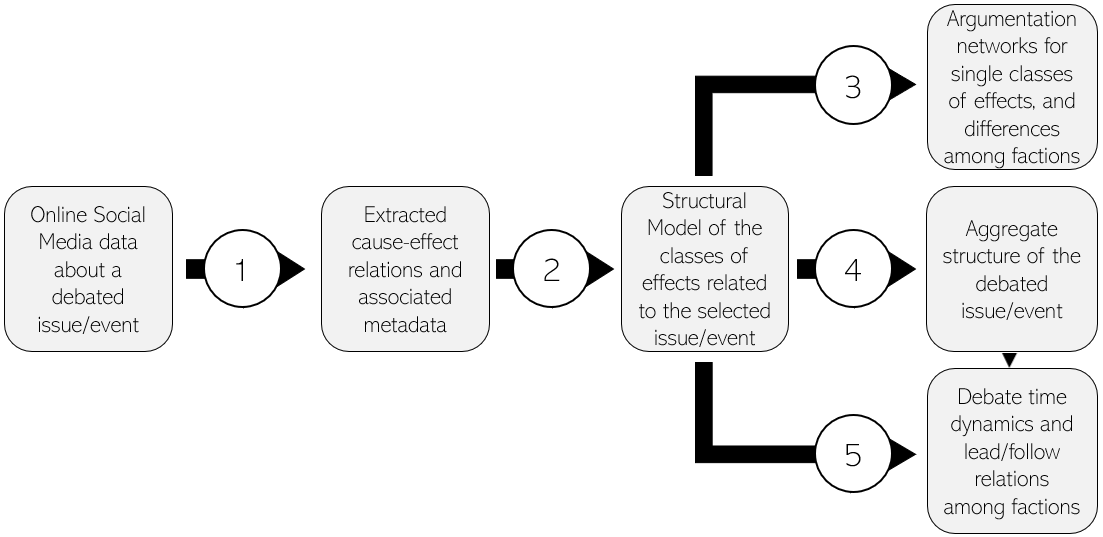}
\end{tabular}
\caption{Summary of the inputs and outputs of the different steps of the analysis}
\label{fig:statistical_rhetorics}
\end{figure*}
\smallskip\par

In this paper, we introduce a methodology for analysing the argumentation patterns and structural properties of online debates and demonstrate its use by analyzing the no-deal Brexit controversy. 
Our fundamental assumption is that arguments do not come insulated in a debate, nor they can be analyzed assuming that they are independent. They are fundamental components of a debate architecture and have to be understood in their interaction with such structure. Debates are about issues that are discussed in specific communication arenas to persuade an audience, orient opinions or decisions. Issues in debates always incorporate a controversial component, associated with partisan factions that support specific views or resolutions. As debates are meant to be persuasive, they imply arguments and counterarguments. Arguments do not come alone but display different degrees of cohesiveness - they are correlated and assembled in coherent blocks. Such correlations provide structure and composite arguments to the debate. Arguments also have internal structure and are expressed through a multitude of semantic components that offer different nuances and interpretations – these are often related to the faction expressing them. Finally, debates happen in time, and are characterized by interaction dynamics among participant factions, with leading and following relations, agenda-setting tentatives, attacks and defenses, shifts of dominant topics and opinions. These can be related both to the endogenous dynamics of the debate and to external events that can steer them. Online debates are no exception:  they often involve a very large number of participants and offer remarkable opportunities to be observed and analyzed over long periods. To apprehend them, it is important to develop a coherent framework of observations and analysis to reveal how the different architectural components of a debate are composed and linked.

\subsection*{\label{sec:intro_rhetoric}{An integrated roadmap to analysing online debates}}

\begin{figure*}
 \scriptsize
{MAY GOV. II \hspace{0.5em} $1^{ST}$ EXTEN. \hspace{0.1em} EU ELECT.\hspace{2em} JOHNSON GOV. I \hspace{4.5em} $2^{ND}$ EXTEN. \& JOHNSON GOV. II }
{\centering
    \begin{tabular}{c}
\includegraphics[trim= 0mm 0mm 22mm 7mm,clip, width= \textwidth]{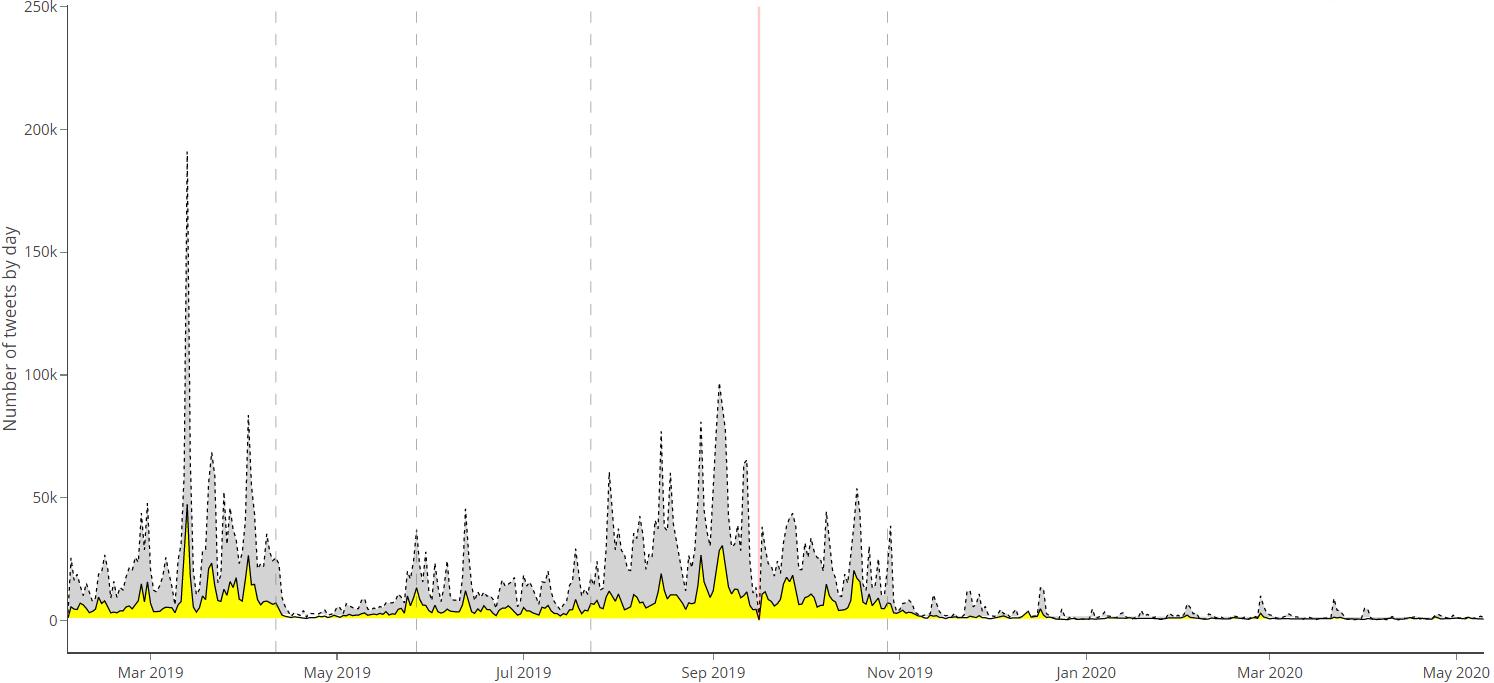}\\
Date\\
\textit{ In {\color{yellow}yellow} tweets counts by day.}\\
\textit{ In {\color{gray}gray} retweets counts by day.}\\
\textit{The {\color{red}red} shaded area represents the day 16 Sept. 2019 that was characterized by technical problems in the client side}\\
\end{tabular}

\caption{Counts per day of tweets and retweets about the ``no-deal''}
\label{fig:all_tweets_counts}}

\end{figure*}

The last decade has been characterised by a sharp rise of interest concerning the use of computational methods for the automated analysis of online debates\cite{barron2018individuals}. This rising interest in the subject has been accompanied by the rapid development of text mining and machine learning methods\cite{blei2012probabilistic}.  
Here we develop a hybrid framework to analyze arguments in online debates. The proposed framework builds on front edge literature by combining and extending in innovative ways existing statistical, text mining and network analysis methods. This, to offer a more consistent and systemic characterization and interpretation of debates which existing tools cannot capture, and to tie together the macro-level (argumentative) and micro-level (phrasal) features of debates. In a way, our approach moves the first steps into the statistical rhetoric of (online) debates. 
In brief, our approach can be summarized in five steps: 
\begin{enumerate}[label=\textbf{\arabic*} -]
    \item\textbf{Argument extraction.} In this paper we focus on causal arguments. Most conventional approaches\cite{willaert2020building} rely on Part-Of-Speech recognition, Relation Extraction and exploit (potentially ambiguous) causal connectives to capture causal statements. Instead, we focus only on verbs as causal markers which express unambiguously the semantics of causation, and for which cause-effect relations can be identified and extracted using simple regular expressions (RegEx).
    Besides being more robust for the type of OSM data employed, this method offers a rich set of possibilities in differentiating types of causal (as well as other modal) arguments. 
    \item \textbf{Aggregating causal arguments in classes (of effects).} The second step consists in aggregating arguments via Structural Topic Modeling (STM) to obtain a limited number of them. In this paper, we focus on the effects which pertain to a single cause/event (i.e. the no-deal). By exploiting the metadata of the tweet and that of the extracted causal relations, we show differences within and between single arguments in terms of factional characterization and types of causal relations employed. 
    \item  \textbf{Comparing faction rhetoric and phrases.} We analyze through an innovative method the internal structure of each argument. We transform each topic into an oriented, weighted graph of words and their associations using uni- and bigrams distributions. This provides considerable additional information on how words are used and phrases are constructed inside the topic. For example, this enables us to see how analogous arguments (e.g. the economic effects of the no-deal) are differently articulated by different factions. 
    \item \textbf{Mapping the structure of the debate.} Arguments are correlated. We filter the network of arguments correlations to uncover the ‘building blocks’ of argumentation and characterize their relationships in terms of types of causal verbs, factions, and inhibition/activation relationships among arguments. 
    
    \item \textbf{Identifying lead and follow faction dynamics.} We look at the time series of arguments proportions to explore their evolution and identify debate leader/follower dynamics among partisan factions taking part in the debate.
    
\end{enumerate}

The output of each step is represented in Figure \ref{fig:statistical_rhetorics}.

 We focus on causal relations as a prominent example of argumentation, but our approach can be extended to other types of relation -such as permission, possibility and influence- used in argumentation\cite{talmy1988force}.

 %for analysing the rhetorics of online debates from a structural perspective

\subsection*{\label{sec:intro_nodeal}{The case of a No-Deal Brexit}}

In this paper, we implement the proposed methodology to analyze the anatomy of the online debate on the ``no-deal'' hard-Brexit by extracting arguments, uncovering their correlation structure, and analyzing the semantics of different factions’ arguments. 
The Brexit online debate is an appealing  case to investigate with the proposed methodology, because, since the 2016 referendum, the ``no-deal'' has polarized public opinions and received great media attention, both offline and online. 
Besides, the attitude towards the ``no-deal'' can be considered among the most polarizing dimensions of the Brexit debate, and is certainly one of the drivers of the 2019 general UK elections results.

For analysing the ``no-deal'' Brexit online debate we use Twitter data in English published from February 2019 to May 2020, which are directly referring to the ``no-deal''  scenario\footnote{See methods Section \ref{sec:method:data_coll} for details.}. 
The time frame of this work includes the first and second extension granted by the EU, covering the period in which the debate has been more active and intense.
By applying our methodology to Twitter posts about the ``no-deal'', we can identify and map online arguments about the expected effects of a hard Brexit and to understand in which terms opposing partisan factions, i.e. \textit{Brexiteers} and \textit{Remainers}, confront each other and try to influence non-partisan online public through distinct argumentation and persuasion strategies.

\begin{figure*}
\scriptsize
{\hspace{-2em}MAY GOV. II \hspace{0.1em} $1^{ST}$ EXTEN. \hspace{0.1em} EU ELECT.\hspace{2em} JOHNSON GOV. I \hspace{4.5em} $2^{ND}$ EXTEN. \& JOHNSON GOV. II }
{
\centering
    \begin{tabular}{c}

\includegraphics[trim= 4mm 0mm 33mm 8mm,clip, width= \textwidth, height=7cm]{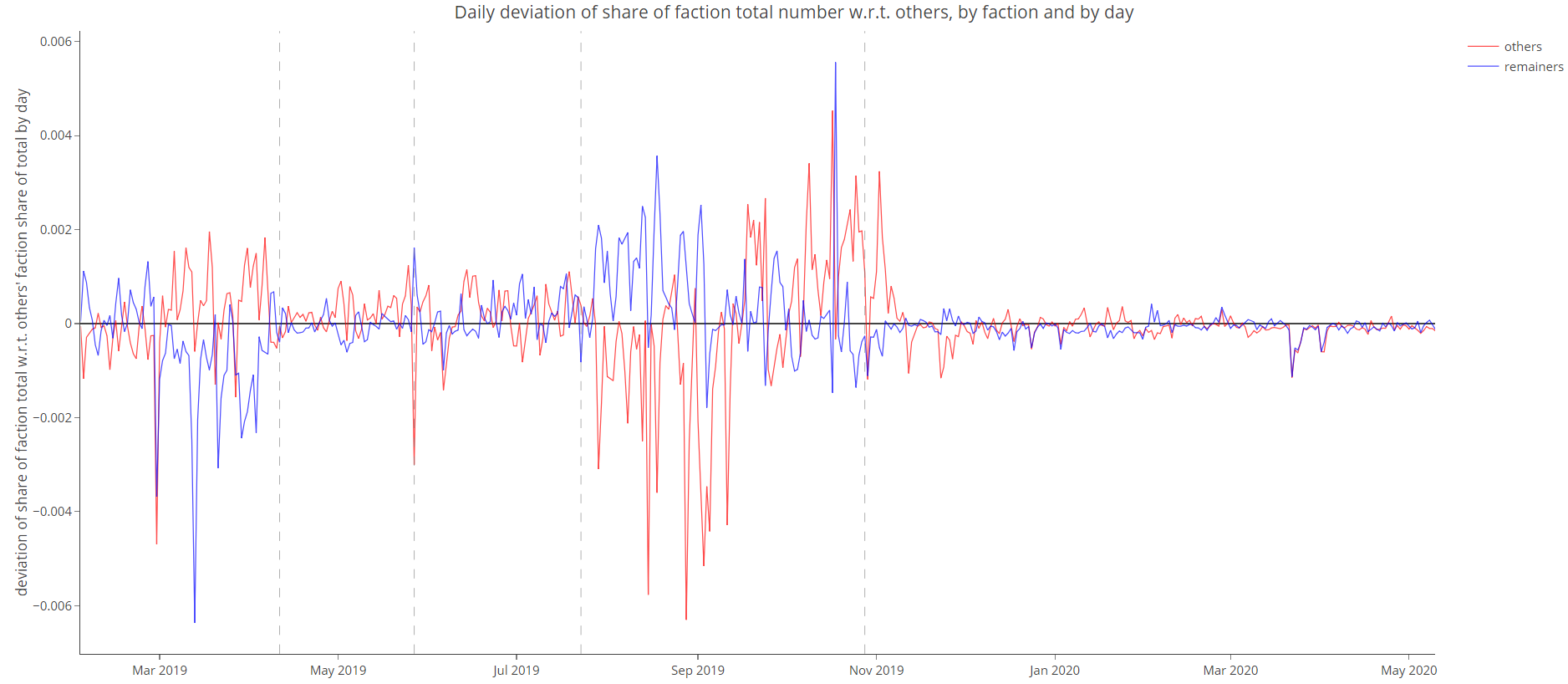} \\
Date\\
\textit{ In {\color{red}red} \textit{Brexiteers}' faction daily tweeting share deviation.} \\ \textit{In {\color{blue}blue} \textit{Remainers}' faction daily tweeting share deviation}\\

\end{tabular}

\caption{Deviation of share of factions' total number of tweets  about ``no-deal''  with respect to \textit{Others}' faction share of total number of tweets, by day}
\label{fig:all_faction_counts}
}
\end{figure*}
 
\section{RESULTS}

\subsection*{\label{sec:nodeal_debate}The no-deal debate on Twitter}

%We collect 9 million Twitter posts mentioning the no-deal, and, 
As preliminary step, using RegEx conditions we identify self-declared Brexit faction partisans among users who tweeted about a no-deal Brexit\footnote{See methods Section \ref{sec:method:rel_extraction} for details.}.
Whereas, users whose Twitter profile \textit{Bio} don't match neither partisan faction (\textit{Brexiteer} and \textit{Remainer}) RegEx conditions are considered part of a residual group, called \textit{Others}. Table \ref{tab:counts} in the Appendix reports the number of tweets and retweets referring to the no-deal by faction. We observe that both partisan factions represent around 2\% of the total volume of activity. Interestingly, \textit{Remainers} exhibit a higher retweet share (79,7\%) with respect to \textit{Brexiteers} (76,2\%) and \textit{Others} (75,2\%). 

The dynamics of no-deal tweets counts by day (see Figure\ref{fig:all_tweets_counts}) allow to single out five stages of activity associated with the phases of the Brexit process.
The first stage ends in mid-April 2019, with the  approval of the EU first “flexible” extension of the UK’s membership. This period is characterized by high volatility and extreme activity peaks, which generally last less than a week. From the end of April to the end of May there is very low activity – with a rapid increase in the week before the EU Parliamentary elections. The third phase, with mid-low volumes of activity and some peaks, follows the success of the Brexit Party at the EU elections and ends with the nomination of Boris Johnson as Tory leader. The fourth and highly active phase corresponds to the first Johnson government and ends with the 2nd extension accorded by the EU. Finally, the subsequent period is characterized by extremely low volumes of tweeting activity about the no-deal. \\
These phases are marked not only by different volumes of tweets but also by different levels of activity by partisan factions with respect to \textit{Others} (see Figure \ref{fig:all_faction_counts}).

The first phase is characterized by a higher initiative by \textit{Brexiteers}, while phase 4 sees an initial burst of activity by \textit{Remainers} followed by a more balanced debate activity of the two partisan factions. 

\begin{table}
\caption{Examples of tweets about the ``no-deal''  containing a cause-effect relation, followed by extracted cause-effect relation and associated metadata}
\centering
\scriptsize
\frame{
    \begin{tabular}{c|c|c|c}
 \multicolumn{4}{l}{Example I} \\
\multicolumn{4}{l}{\includegraphics[trim= 0mm 0mm 0mm 0mm,clip, width= 0.5\textwidth]{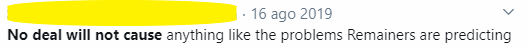}} \\
\textit{cause-side} & \textit{effect-side} & \textit{rel.type} & \textit{negated}\\\colrule
No Deal & anything [..] predicting & Causation & TRUE\\\hline
 \multicolumn{4}{l}{  } \\
 \multicolumn{4}{l}{Example II} \\
\multicolumn{4}{l}{\includegraphics[trim= 0mm 0mm 0mm 0mm,clip, width= 0.5\textwidth]{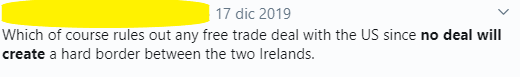}} \\
\textit{cause-side} & \textit{effect-side} & \textit{rel.type} & \textit{negated}\\\colrule
Which [...] no deal & a hard [...] Irelands. & Creation & FALSE\\
\end{tabular}
}
\label{fig:causal_tweets}

\end{table}

%Most important to understand the nature and rhetorical style of the debate is to look to if and how different factions use causal markers in their tweets and retweets  (see Table \ref{tab:relationcounts} in Appendix). 

Our approach to argument extraction is based on verbs\footnote{See methods Section \ref{sec:method:rel_extraction} for details.}.
Remarkably \textit{Remainers} use in their tweets (almost 28\%) more causal verbs with respect to \textit{Brexiteers}. This difference is attenuated, but persists, for retweets (see Table \ref{tab:relationcounts_appendix} in the Appendix). It is possible to capture semantic differences among causal arguments. In particular, we focus on three groups of causal verbs. Two groups clearly express the polarity of the causation relation, we label them as \textit{Destruction} and \textit{Construction} verbs. Whereas, the third group, which has a neutral polarity, is simply called \textit{ Causation}.
By applying our relation extraction method we obtain for each group of verbs ordered pairs of causes and effects, with the associated metadata (see examples in table \ref{fig:causal_tweets}).

We have 204 648 relations, of which 36.116 contain ``no-deal''  inside the cause side of the relation. From now onward we analyse only this set of relations.
 Further argumentation style differences among factions exist. In particular, \textit{Brexiteers} put stronger emphasis than \textit{Remainers} on \textit{Creation} relationships, while the converse is true for \textit{Destruction} relations (see Table \ref{tab:relationcounts} in Appendix). This is true also when we consider only the relations which are not negated (see table \ref{tab:relationcounts_appendix} in Appendix).

\begin{table*}
\centering{
\caption{
\label{tab:summary_table_top10}%
Summary table of top 10 topics  (by overall topic proportion), with top 10 tokens (by token probability) by topic and by faction. Row color scale represents significant (at the 0.01 significance level) differences of the estimated topic proportion coefficients of the two partisan factions. The closer is the row color to the corresponding faction color, the more characterizing is a specific topic for one of the two partisan factions.  Characterizing topics for \textit{Remainers} are in {\color{blue} blue}, whereas characterizing topics for \textit{Brexiteers} are in {\color{red} red}. For the full list of topics we refer to Table 2 in the Supplement.
}
\includegraphics[trim= 0mm 0.4mm 0mm 0mm,clip, width= 1.03\textwidth]{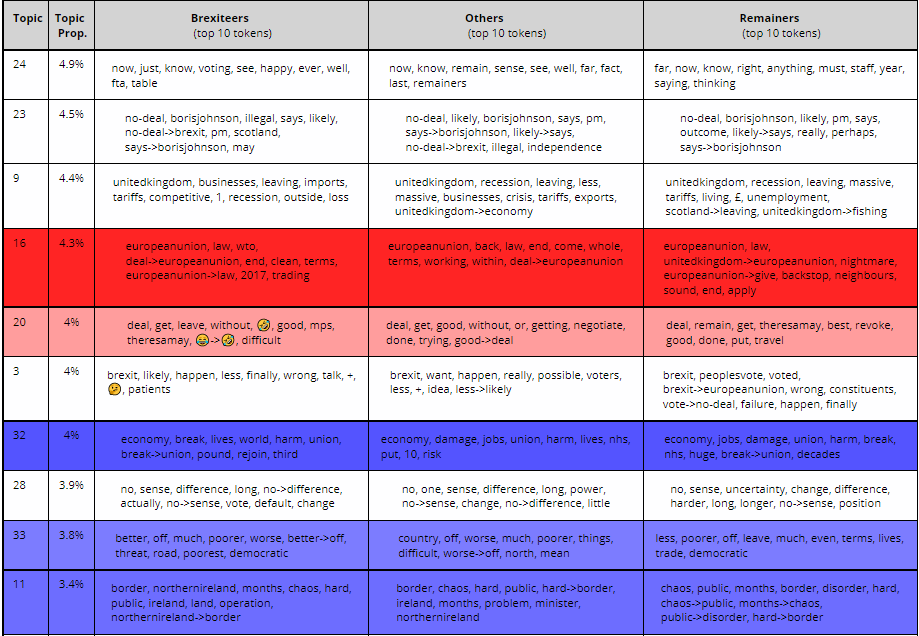}

}
\end{table*}

\subsection*{\label{sec:nodeal_aggregation}Aggregating arguments in classes (of effects)}

% \begin{enumerate}
% \item briefly present the outcome of the STM model with standard tools;
% \item present topics as topics and show which ones are more discussed by each faction;
% \end{enumerate}

Since extracted relations include a large and varied population of effects, there is a need to aggregate the phrases describing the effects of the no-deal in a manageable number of effect topics.
We do so through a STM applied on the effect side of extracted relations.
One major advantage of using a STM, is that it allows to include metadata as covariates affecting topic proportions and topic contents. This makes it possible to capture, for example, differences in 
how (topic contents) and how much (topic proportions) factions speak about the different expected effects of the no-deal. These differences cannot be captured with classical topic modeling techniques, like LDA or CTM.
The covariates that we allow to affect topic proportions are time ($t$), the user faction ($fct$), the relation verb negation ($neg$) and the relation verb group ($rel.type$). By so doing, the model can be used to identify which topics characterise each partisan faction, i.e. topics that are more likely to be observed conditionally on the user belonging to one partisan faction with respect to the other. 
For explaining the topic content we consider only the user faction ($fct$). This allows to capture differences between groups, in terms of how they debate about each topic, by giving different probability weight to words (unigrams) and collocations (bigrams) that are used to express causal beliefs about the expected effects of a no-deal scenario. Table  \ref{tab:summary_table_top10} shows the 10 most probable tokens (unigrams and bigrams) for the top 10 topics, ranked in decreasing order by overall topic proportion. 
For example, topic 16, concerning trade agreements,  (which is more likely used by \textit{Brexiteers}) clearly shows differences in the evaluation of \textit{Brexiteers} (stressing clean terms and the opportunities to use WTO trading agreements) and Remianers (which see the same issue as a nightmare). In topic 32 (which is characterizing \textit{Brexiteers}), which is about the economic consequences of the no deal, \textit{Remainers} stress damages to jobs and the stress on the health system ("nhs" token), which disappear from  the top list of words for the \textit{Brexiteers}.
Interestingly, for topic 28, which is characterizing neither partisan faction, that focuses on the sense (and non sense) of a no-deal scenario, the \textit{Brexiteers} and \textit{Others} appear to claim that a no-deal will likely produce no difference (see rank of bigram token: "\textit{no} $\rightarrow$ \textit{difference}"), whereas for \textit{Remainers} "\textit{uncertainty}" is a more high-ranked effect.
Finally, \textit{Remainers} appear to be more concerned than \textit{Brexiteers} and \textit{Others} by the scenario of Scotland leaving UK as a result of a no-deal (see topic 9).

\subsection*{\label{sec:nodeal_aggregation} Exploring faction rhetoric and narratives of no-deal arguments}

\begin{figure*}
    \centering
    \hspace{-1em}
    \begin{tabular}{c|c}
 \multicolumn{2}{c}{\includegraphics[trim= 1.25mm 0mm 0.25mm 1mm,clip, width= 16.5cm]{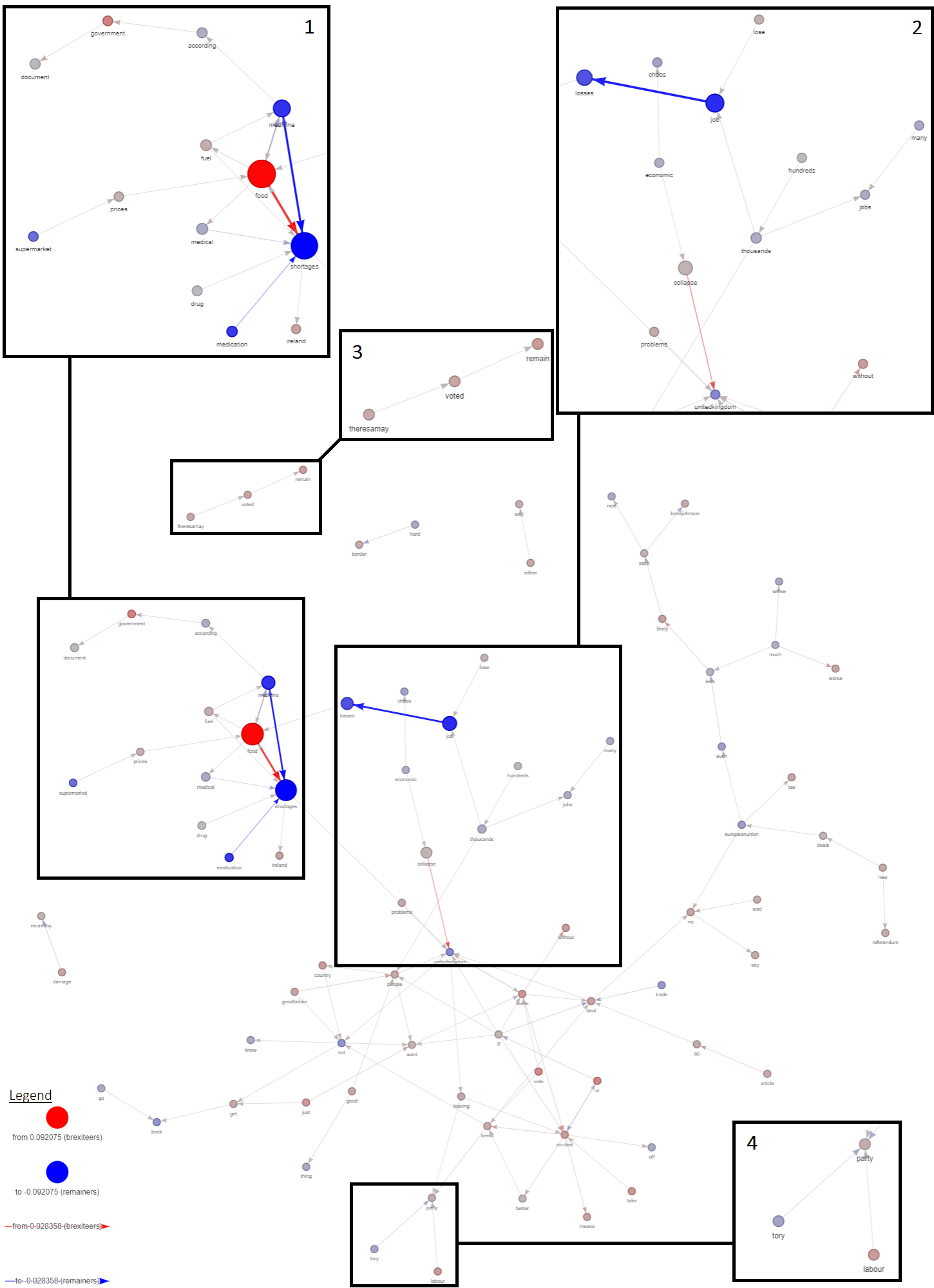}}
\\
\end{tabular}
    \caption{Topic 2 argumentation network.  Unigrams (nodes) and bigrams (edges) have been filtered ($80^{th}$ percentile threshold) on their probabilities for the non partisan faction \textit{Others}. Node size and edge width represent probabilities of unigrams and bigrams for the non partisan faction \textit{Others}. Node/edge color scales are used to represent the differences in unigram/bigram probabilities between partisan factions (\textit{prob.} \textit{Brexiteers} \textit{minus} \textit{prob.} \textit{Remainers}). Numbered boxes display zoomed areas of interest.\\}
    \label{fig:topiczoom}
\end{figure*}

Figure \ref{fig:topiczoom}, shows the argumentative network of topic 2, which has been reconstructed using estimated token probabilities for that topic\footnote{unigrams' probability is used to weight nodes and bigrams' probability is used to weight edges for a selected topic}.\par
To see in which terms the two partisan factions intervene in the non-partisan debate concerning topic 2, we create the topic network for the faction \textit{Others} and filter separately nodes and edges keeping only the 80th percentile, to prune the network from terms (unigrams) and collocations (bigrams) less frequently employed by non-self-declared partisan users within this topic. We subsequently overlay to this network\footnote{using a red-gray-blue color scale, where gray is centered at 0, i.e. no difference between faction probabilities for that token (unigram/bigram)}, the differences in token probabilities between the \textit{Brexiteers} and \textit{Remainers}, for each token in the former network. The color of edges and nodes hence represents the partisan faction by which a specific term or collocation is more employed in relation to the selected topic\footnote{See methods Section \ref{sec:method:intra_topic} for details.}.\par

The terms "\textit{shortages}", "\textit{food}" and "\textit{medicine}" are among the most relevant terms for non partisan users (see box n.1 figure \ref{fig:topiczoom}). The probabilities of these terms and collocations are very different for \textit{Brexiteers} and \textit{Remainers}. For example, while \textit{Brexiteers} focus more on the term ``\textit{food}" and the collocation ``\textit{food }$\rightarrow$ \textit{shortages}", \textit{Remainers} focus relatively more on the terms ``\textit{medicine}", ``\textit{medication}", and, the collocations  ``\textit{medicine} $\rightarrow$ \textit{shortages}" and ``\textit{medication} $\rightarrow$ \textit{shortages}".\par
The \textit{Remainers} also tend to focus more on the possible effects on unemployment of a no-deal scenario (see box n.2), both the unigrams ``\textit{job}" and "\textit{losses}", and the bigram ``\textit{job} $\rightarrow$ \textit{losses}" have higher probability for \textit{Remainers} w.r.t \textit{Brexiteers}. 
Finally, box n. 4 shows that members of partisan factions are more incline to talk about the party representing the other faction with respect to their own: \textit{Remainers} use relatively more frequently ``\textit{tory}" and ``\textit{tory} $\rightarrow$ \textit{party}", whereas \textit{Brexiteers }  use more frequently ``\textit{labour}" and ``\textit{labour} $\rightarrow$ \textit{party}". This signals that while debating about the effects of no-deal, partisan faction members also argue about the no-deal narratives employed by the opposing faction.

\pagebreak
 \subsection*{\label{sec:nodeal_aggregation} The structure of the no-deal debate}
\begin{figure*}

\centering{
\includegraphics[trim= 0mm 0mm 0mm 0mm,clip, width= 0.75\textwidth]{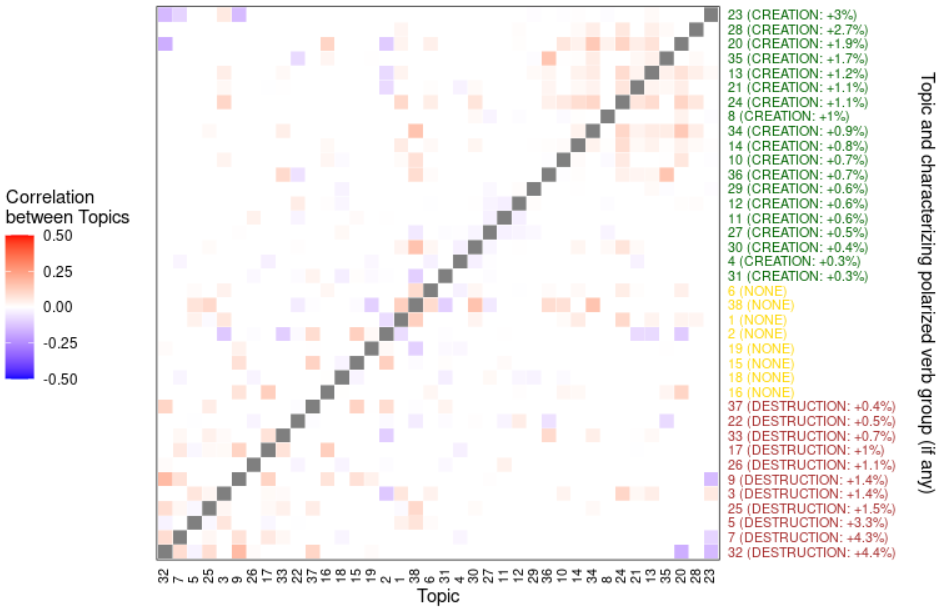}
\caption{Heat-map of correlations between topics, in {\color{red}red} positively correlated topics, in {\color{blue}blue} negatively correlated topics. 
Non significant correlations have been identified (and set to 0) using the method proposed by Meinshausen and Buhlmann\cite{meinshausen2006high}.
The topics have been reordered on the basis of the estimated difference between the coefficients of polarized verb groups' (\textit{Creation} Vs \textit{Destruction})  effects on topic proportions ($\hat{\beta}_{k,rel=Destruction}-\hat{\beta}_{k,rel=Creation}$). If this difference is significant at the 0.01 level, the label ({\color[rgb]{0.0, 0.5, 0.0} \textit{Creation}} or {\color[rgb]{0.47, 0.27, 0.23}\textit{Destruction}}) of the verb group characterizing the topic is contained in brackets after the topic number, followed by the value of the estimated difference (sign adjusted).  If non significant, the label {\color{yellow}\textit{None}} follows the topic number (in brackets).}
\label{fig:heatmap_correlations}}
\end{figure*} 

\begin{figure*}
    \centering
    \begin{tabular}{c|c}
% \hspace{0pt} Filtration threshold=0.20  &  Filtration threshold=0.17 \\
%\includegraphics[trim= 0mm 0mm 0mm 0mm,clip, width= 3.5cm]{figures/corr_020.PNG} & 
%\includegraphics[trim= 0mm 0mm 0mm 0mm,clip, width= 4cm]{figures/corr_017.PNG} \\\hline
 Filtration threshold=0.14  & Filtration threshold=0.10\\
\includegraphics[trim= 0mm 0mm 0mm 0mm,clip,width= 5.5cm]{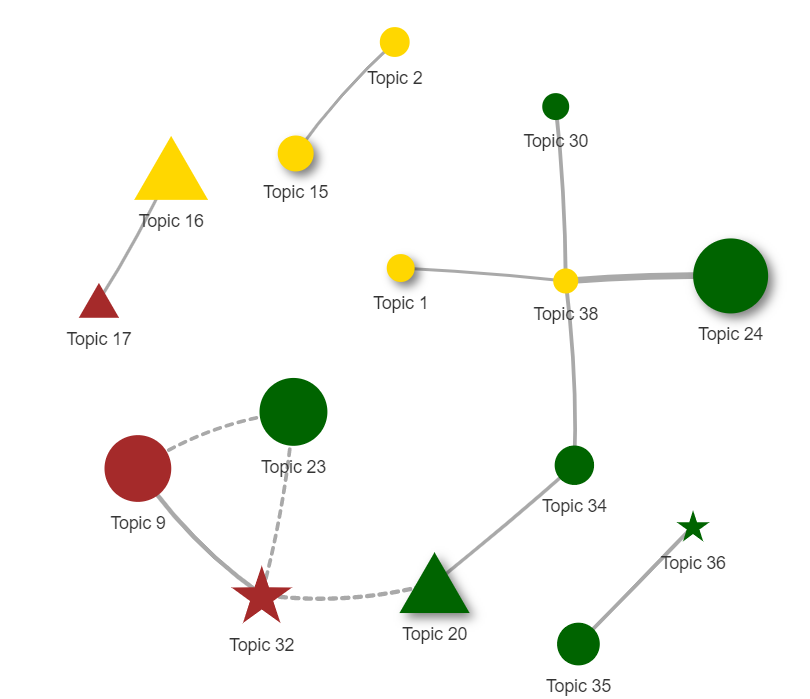} & \includegraphics[trim= 0mm 0mm 0mm 0mm,clip,width= 6.5cm]{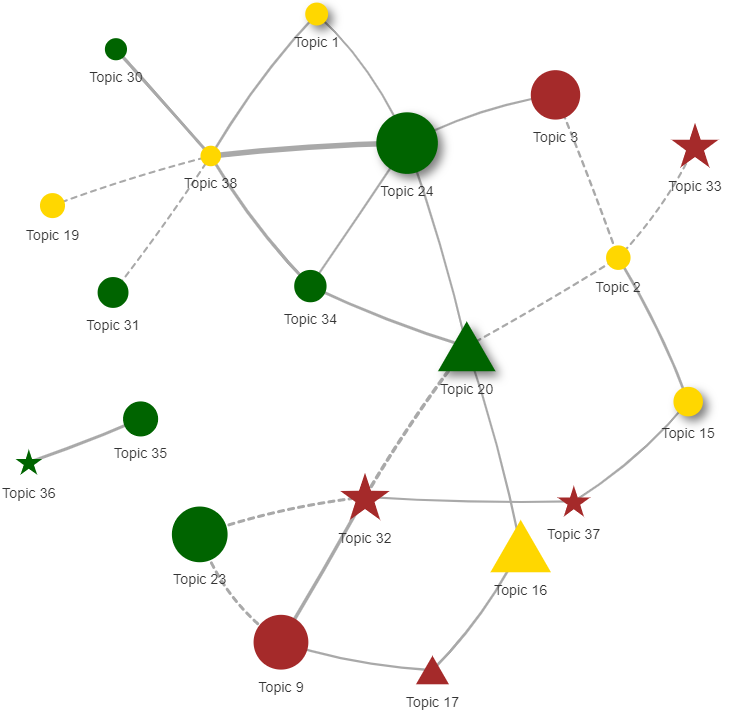}
\\\hline
 \multicolumn{2}{c}{Filtration threshold=0.05}\\
 \multicolumn{2}{c}{
\includegraphics[trim= 0mm 0mm 0mm 0mm,clip,width= 10cm]{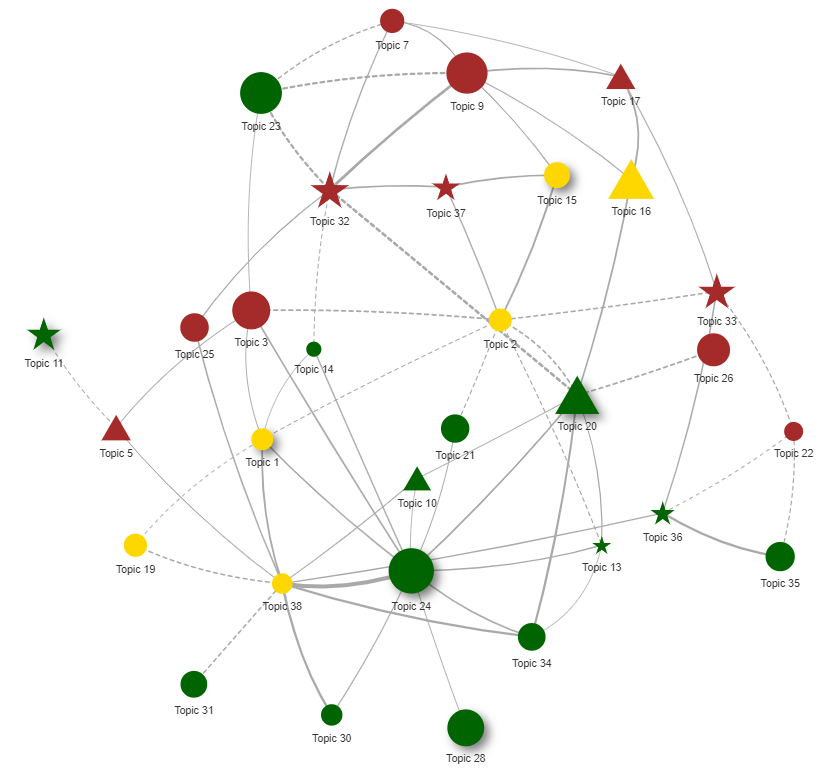} \includegraphics[trim= 0mm 0mm 0mm 0mm,clip,width= 4cm]{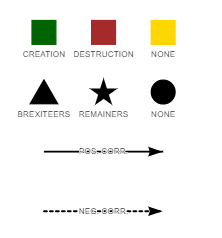}}
\\
%\begin{rotate}{90} \hspace{30pt} Legend\end{rotate}
\end{tabular}
    \caption{Filtered topics correlation network (absolute value) and faction/type covariates. Node color represents predominant partisan (\textit{Brexiteer}/\textit{Remainer}) faction (p.value$<$0.01), see legend for color details.\\
Node shape represents predominant \textit{Destruction}/\textit{Creation} relation type (p.value$<$0.01), see legend for shape details.
Node shadow if negated relation is predominant (p.value$<$0.01). Solid lines represent positive correlations whose absolute value is higher or equal to the filtration threshold. Dashed lines represent negative correlations.}
    \label{fig:filtration}
\end{figure*}

In a debate, classes of effects (i.e., topics) related to a common cause are not independent.  
Some topics are likely occurring together in sentences, whereas others mutually inhibit each other. For example, some effects are semantically related through the verbs expressing causal relations.  Topics tend to be clustered according to verbs' polarity. Effects related to \textit{Destruction} tend to be positively correlated among them and negatively correlated with those related to \textit{Creation}. The same is true for \textit{Creation} verbs. Figure \ref{fig:heatmap_correlations}, displaying the correlations among topics, shows the near-decomposability of causal arguments in blocks around the two diagonals of the matrix.
One can capture a finer-grained structure of the debate by filtering the correlation network \footnote{See methods section \ref{sec:method:aggregatedebate} for details}. Filtration creates a continuum of networks resulting from the deletion of edges whose weight is below a given threshold varying over the range of observed weights. In our setting weights represent the value of correlations, and these are filtered according to their absolute values. At higher levels of the threshold 
the network displays strong relations between arguments. The positively connected parts of the graph capture the conceptual building blocks of the debate's arguments, i.e. topic constellations. As the threshold is relaxed, a weaker (but still significant) set of relations among topics emerges and "assembles" the building blocks (see Figure \ref{fig:filtration}). 
% emergence of topological structures

Additional structural information is gained by displaying significant topic covariates in a topic correlation network, by typing the nodes by dominant partisan faction (shape),  dominant polar verb group (color), and negation of the relation (shadow), whenever the difference among types is significant\footnote{See Section 1 in the Supplement.}.
Filtration of such typed network allows highlights that the topic correlation network exhibits type-based assortativity, i.e. assortative-mixing of the topics for multiple (simple and composite) typing dimensions. \par
Figure \ref{fig:filtration} shows clear evidence of polar group verb assortativity (up to the 0.14 threshold),  and partisan faction assortativity (up to the 0.10 threshold)\footnote{We remark that the only pair of \textit{Creation}-\textit{Destruction} topics that is positively correlated at the filtration level 0.10 is \{topic 24, topic 38\}. This pair of topics contains topic 24 that is one of the few topics for which the negation has a significant positive effect on the topic's expected proportion, i.e. the topic is more likely to appear conditional on the presence of a negation of the relation's verb phrase (e.g. \textit{"[No deal] does not create [topic 24]"}).}.

Moreover, up to the 0.10 threshold, positively correlated triangles (3-cliques with non dashed edges) are formed by topics of non opposing types. These positively correlated triangles are also coherent in terms of polar faction types, being made either by topics that are not characterizing any of the partisan factions or by topics that are not of opposing factions.\par
Up to the 0.10 threshold, the only triangle containing both positively and negatively correlated topics is \{topic 23, topic 32, topic 9\}, which is balanced and coherent being formed by two positively correlated \textit{Creation} topics(32 and 9), that are both negatively correlated with the third one that is a \textit{Destruction} topic (23). \par
More generally, as we lower the threshold to 0.05, the connected components of the graph grow respecting a basic triangle balance principle, i.e. connected triplets of topics are always made of an odd number of positive edges.
At thresholds below 0.05 most of the triangles are still balanced, but, as the threshold further lowers some imbalanced triangles appear, maybe displaying the appearance of “relational noise”.\par

 \subsection*{\label{sec:nodeal_aggregation} The lead and follow dynamics of the no-deal debate}
 
 \begin{figure*}
\centering{
\includegraphics[trim= 0mm 0mm 0mm 2mm,clip, width= 1.03\textwidth]{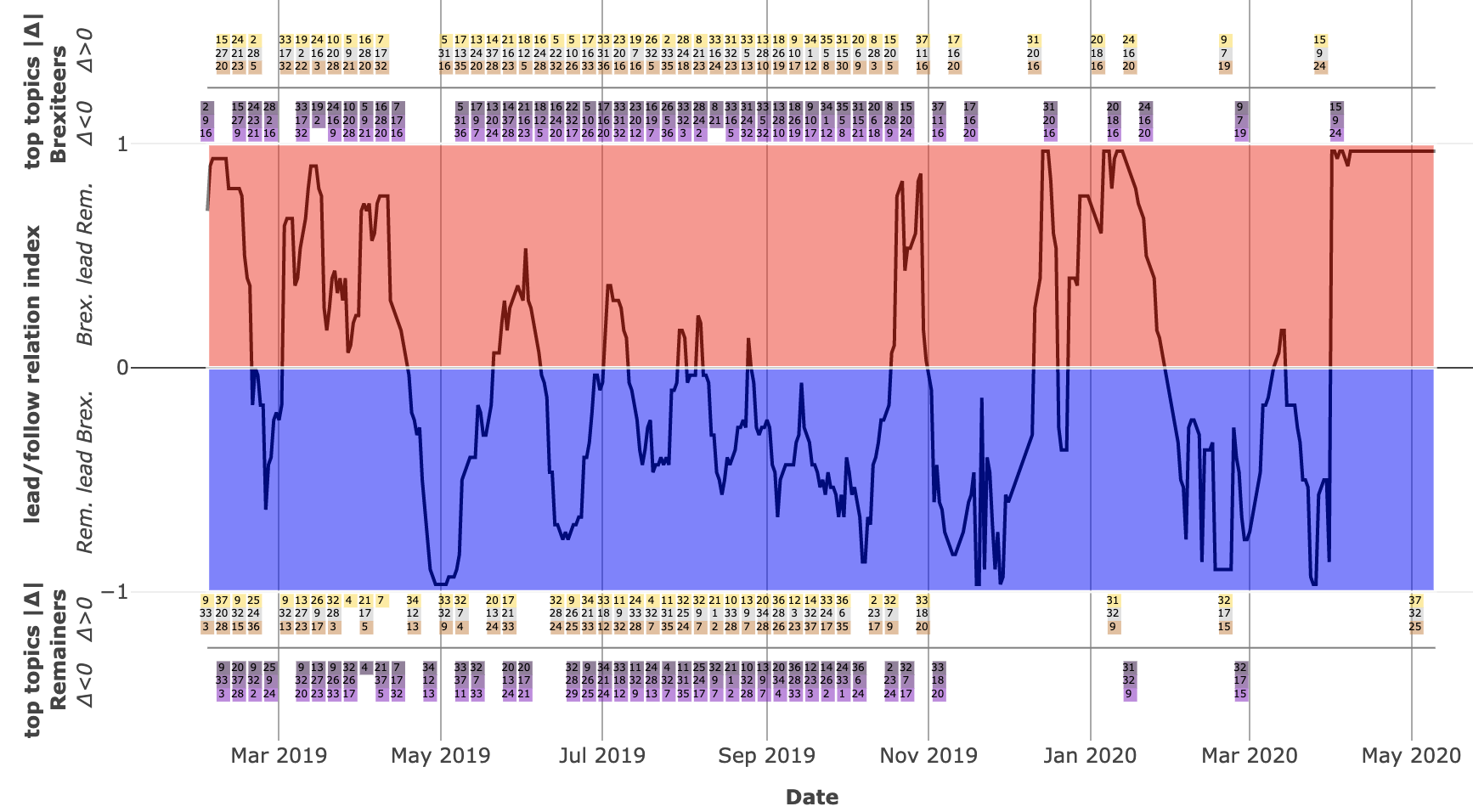}
\caption{\textit{Central plot:} lead/follow relation index among partisan factions. The index has been constructed using the method proposed by \cite{amornbunchornvej2018framework},  it is based on Dynamic Time Warping of the 38-dimensional daily topics proportions series by faction, using a time window of 30 days, a max lag window of 6 days, and a window shift of 1 day. The closer the index value is to 1 the more intense is the lead of \textit{Brexiteers} on \textit{Remainers} on that day, the converse is true for values close to -1.\\ \textit{Upper and lower plots:} Top topics by absolute weekly topic-proportion variation\footnote{$|w_{x,k,i}-w_{x-7,k,i}|$ see methods section \ref{sec:method:lead-follow} for details}. Inside gold, silver and bronze boxes are contained the IDs , of the top 3 topics, in terms of largest weekly positive variations, representing respectively the 1st, 2nd and 3rd ranked topics. Inside the dark violet, violet and light violet boxes are contained the IDs of the top 3 topics in terms of largest weekly negative variations, representing respectively the 1st, 2nd and 3rd ranked topics. Some boxes may be missing when for less than 3 topics non-null (positive/negative) weekly topic proportions changes are observed. 
}
\label{fig:lead_follow}}
\end{figure*}

%\begin{enumerate}
%\item Time dimension of the debate, which factions are synchronized? which lead/follow?
%\item following the crowd? or following the factions?
%\end{enumerate}

 Debate's lead and follow relations among partisan factions are important because they reveal the capacity of a faction of setting the thematic agenda for a specific debate in a specific moment in time. Leader-follower relations among factions can be seen as processes of highly coordinated debate activity: when a faction starts discussing more or less about specific effect topics the other faction then follows similar topic proportion variations. In our setting, a leading faction can be seen as a synchronized group of partisan users initiating a debate shift, whereas a following faction can be seen as a rival group of partisan users pushed to respond to the other faction by discussing in the following days the topics proposed by the latter.

To identify possibly dynamic lead-follow relations among \textit{Brexiteers} and Remainers, we apply a Dynamic Time Warping\cite{berndt1994using} method called FLICA\cite{amornbunchornvej2018framework}. This method allows us to infer time-varying lead-follow relations between pairs of multidimensional time series, in our case between the faction-specific 38-dimensional topic proportion series, each dimension representing the average daily proportion of a specific topic for a specific partisan faction\footnote{We use a window of 30 days, a max lag window of 6 days, and a window time shift of 1 day. Our results are robust to changes of the three parameters, in particular to the  max lead/lag window which is the most relevant parameter representing the max range of the warping. With max warping window values from 3 to 10 days, lead/follow patterns remain consistent. While, as expected, reducing the warping range mostly affects the amplitude of the observed oscillations of the index.}.

As we can see from the central plot in Figure \ref{fig:lead_follow}, which shows the lead/follow index obtained by applying FLICA, lead-follow relations among the two partisan factions exhibit multiple cycles with different intensities and degrees of persistence.  In particular, by relating the peaks and troughs of the index to political events in the UK, we notice that peaks correspond to moments in which politicians and political parties close to \textit{Brexiteers} were in a position of strength, whereas the converse is true for troughs. Interestingly, rapid shifts in leadership occur after the following events:
\begin{itemize}
\item  \textbf{MPs Amber Rudd ultimatum (22 Feb. 2019)}: \textit{Brexiteers} lead the debate in the first two decades of February, then \textit{Remainers} take the lead until the beginning of March 2019;
\item  \textbf{1st extension (10 Apr. 2019)}: \textit{Brexiteers} lead the debate from the beginning of March to mid April, then \textit{Remainers} take the lead until the end of the second decade of May 2019;
\item  \textbf{EU elections (23 May 2019) and the resignation of Theresa May (7 Jun. 2019)}: \textit{Brexiteers} lead the debate from May 21st to the day in which T. May's resignation becomes effective;
    \item  \textbf{Yellowhammer plan leak (18 Aug. 2019)}: \textit{Remainers} lead the debate from mid August until mid October 2019 (excluding 25 Aug. 2019);
    \item  \textbf{2019 United Kingdom General Election (12 Dec. 2019)}: \textit{Remainers} lead the debate from the beginning of November until two days before the GE, then \textit{Brexiteers} take the lead until the end of January 2020 (excluding 21-23 Dec. 2019);
\end{itemize}

Moreover, by jointly analysing the lead/follow index and the weekly topic variations (see upper and lower plots in Fig. \ref{fig:lead_follow}), we observe some extremely interesting patterns. First (i), periods where \textit{Remainers} lead or where leadership starts shifting towards them are often characterised by the presence among the top 3 topics (by weekly positive-sign variation) of topic 32 and topic 9, which are both related to the economic effects of a no-deal scenario. These two topics are also dominated by \textit{Destruction} causal verbs and form one of the strongest macro-argument components identified in Figure \ref{fig:filtration}. Moreover, topic 32 was also found to be characterizing for Remainers.  Second (ii), periods where \textit{Brexiteers} lead or where leadership starts shifting towards them are similarly characterised by the presence among the top 3 topics (by weekly positive-sign variation) of the topic 20 and topic 24, which are both dominated by \textit{Creation} causal verbs, and trivialize the alternative to a no deal scenario by ridiculing the difficulties and delays in the negotiations of a trade deal with the EU. Topic 20 is characterizing for \textit{Brexiteers} and contains many emojis that jokingly refer to the extremely remote possibility of reaching an agreement and hence avoiding a no-deal scenario.

Both points suggest that the structural properties identified in the  effect topic-correlation network analysis also play a role in the dynamics of lead/follow relations among factions – further supporting the importance to consider the interdependence of arguments and of levels analysis.

\section{DISCUSSION}

We have shown that the structure and dynamics of online debates connect arguments in a coherent way different levels and roles. 
Our reconstruction of causal arguments in the No Deal debate allows to unveil how the debate dynamics relate to external events, how the different causal arguments mutually connect, and how they are shaped by factional interests. For example, \textit{Remainers} resort more extensively to causal arguments than \textit{Brexiteers} do, they emphasize the potential destructive causal effects of the no deal Brexit, and stress its (negative) economic implications. 
%Opposing factions talk more about their opposite factions than about themselves – arguments are polemical. We show how arguments correlate among them, by supporting or mutually inhibiting each other (for example, arguments built on destructive verbs tend to be cohesive among them and inhibit constructive ones). 
In particular, we show how the network of arguments displays polarized assortativity around structurally balanced building blocks that aggregate topics by type and factional orientation. Correlated constellations of arguments also play an important role in the debate dynamics, as they often mark, in conjunction with external events, shifts in the factional leadership of the debate.

Thus, better insights can be obtained by addressing the complex architecture of debates through approaches that combine different tools in a coordinated way. In this paper we explore a multi-step approach that traces a methodological roadmap through the different architectural components of the debate – pointing in the direction of the development of  the statistical rethorics of debates. In order to do this, we had to adapt pre-exisiting tools in innovative ways, e.g. by combining structural topic modeling with network modelling.  This allows to ask new questions about the structural properties of debates that generalize beyond the specificity of our case study. For example, the filtration of the correlation network of effect topics suggest some interesting structural regularities – e.g. argument triangles are all structural balanced until very low level of correlation (maybe corresponding to relational noise) are considered.

Besides this general contribution, our paper further contributes to more specific streams of literature. 

%{\color{blue}Cause maps}\\
%{\color{blue}Topic Models}\\
In relation to Structural Topic Models, this work highlights how classification algorithms that jointly exploit text and metadata can be fruitfully used not only with short online social media posts, like Tweets\cite{curry2019may}, but also with subsets of short posts which may represent phrases or causal relations extracted with RegEx or NLP algorithms from the former. Moreover, besides using classical document-level covariates, like author and publishing date, this work shows that one can transform otherwise non-exploited textual data from a post or from its metadata in valuable categorical covariates, like the faction and causal relation verb type. 
%These ad-hoc covariates generated from text can hence be used to evaluate in which terms these dimensions are useful in predicting and explaining topic proportions, topic contents, and more generally for interpreting the aggregate structure of the debate and its topical constellations. Constellations that in this work appear to be subject to assortativity driven by aforementioned faction and (causal) verb type covariates.       

%{\color{blue}Networks}\\
From the point of view of network science, this paper clearly demonstrates that from the micro to the macro scale, graphs rather than sets, appear to be the most appropriate way to analyse and model debates: At the micro level, the inclusion of bigrams in modelling setting allowed us to reconstruct faction-specific argumentation networks, which highlight in which terms single factions intervene in a topical debate; At the macro level, the combination of the topics correlation matrix with predominant covariate-level labels allow us to model the architecture of the debate, and identify its salient dimensions and topological properties.  
From a broader perspective this work contributes to the emerging field of narrative economics\cite{shiller2017narrative} and constructivist approaches to socioeconomic issues\cite{searle1995construction} by offering an extensive framework for studying online causal debates and their dynamics.
%{\color{blue}Limitations}

Our analysis of the No Deal debate considers only causal arguments that have a single cause (the “No Deal” causal factor) and focuses on different effects associated to it. This constraint was introduced to keep our analysis simple enough, but is not a an intrinsic limitation of our approach. Indeed, one may consider multiple causes of  a single effect, or multiple causes of multiple effects as well. 
A second limitation is that we reconstructed the phrasal networks only within each effect topic. However, as we have shown, arguments are correlated and form detectable constellations of topics. It is possible to reconstruct phrasal networks for such constellations as well. This would potentially bring to light more connections among words and the phrasal constituents of more complex arguments.
Thirdly, more systematic information could be extracted by looking at structural indicators of networks, such as centrality, betweenness, modularity and  community structure. We explore such issues in a separate paper\cite{warglien2021}.
Finally, in our analysis we didn't consider retweets which provide natural indicators of the social resonance of arguments.  As this deviates from the main focus of this paper, we defer this analysis to a different one.
%Technically speaking, using retweets to estimate STM models would introduce  redundancies that would affect the quality of the model. For example, we could integrate ex-post the retweet data to understand whether the retweet probability reflects the faction characterization of topics identified with tweets. As this deviates from the main focus of this paper, we defer this analysis to a different one.

\clearpage

\section{METHODS}

The data acquisition, preparation and modelling procedures used in this study
are as follows.

\subsection{Data collection and pre-processing}\label{sec:method:data_coll}
Tweets about the no-deal have been collected through Twitter's Stream API, from the beginning of February 2019 to beginning of May 2020\footnote{Twitter's V1 filtered Stream API endpoint has been queried through DMI-TCAT\cite{borra2014programmed}, using as 
stream filters 'no deal' and 'no-deal'}. In total, more than 9 million tweets were downloaded and archived in a SQL database. 
The downloaded data were then pre-processed to remove tweets that are unrelated to the no-deal Brexit. In particular, tweets related to the US-China trade war (and its no-deal scenario) where identified and hence removed using a RegEx containing references to these two countries and their leaders\footnote{Regex with case sensitive matching: "China$\mid{}$china$\mid{}$u[.]s[.]a$\mid{}$u[.]s$\mid{}$ U[.]S$\mid{}$US$\mid{}$USA$\mid{}$UNITEDSTATES$\mid{}$UnitedStates$\mid{}$unitedstates$\mid{}$UNITED STATES$\mid{}$United States$\mid{}$united states$\mid{}$trump$\mid{}$Trump$\mid{}$TRUMP$\mid{}$XI$\mid{}$Xi"}.
The cleaned data-set contains 9004927 tweets about the no-deal Brexit and their metadata.
%In recent years, the contents of Twitter posts, called tweets, have been used in several applied-fields of research, including political studies, sociology and economics. In relation to economics, tweets have been used to build investor and consumer sentiment measures. On the macroeconomic side, Twitter data has been used to construct economic uncertainty indicators that can explain volatility and investment. Twitter data has also been successfully used for catastrophic event detection and disease surveillance purposes.

\subsection{Arguments and argument-specific covariates extraction}\label{sec:method:rel_extraction}

%\begin{enumerate}
%\item Extracting causal clauses with RegEx
%item brief review of other methods
%\item Present covariates and their construction
%\end{enumerate}

To find and extract from the no-deal Tweets segments of text that identify cause-effect relations we employed a RegEx algorithm that exploits verbs. In particular, a list of verbs and verb phrases related to \textit{Causation}, \textit{Creation} and \textit{Destruction} (see Table 7 in the Supplement) are used to build a set of RegEx functions, which are hence used to identify and isolate the two sides of each cause-effect relation contained in the corpus of tweets. The steps of the the cause-effect extraction process can be summarised as follows\footnote{For RegEx algorithm details and commented code we refer to Table 8 in the Supplement.}:
\begin{enumerate}
    \item[1] Tweets are segmented in sentences using punctuation characters\footnote{the Quanteda library\cite{benoit2018quanteda} was used for this purpose.}.
    \item[2] For each sentence:
    \begin{enumerate}
    \item[2.1] Verb phrases related to cause-effect relations are identified through RegEx functions. 
     \item[2.2]  If there are one or more RegEx matching, for each matching we identify and record the verb phrase's:
      \begin{enumerate}
   \item[2.2.1] position (\textit{start char.} and \textit{end char.});
   \item[2.2.2]  type (\textit{causation}, \textit{creation}, or \textit{destruction});
   \item[2.2.3]  negation (e.g., \textit{''Y will \textbf{not} cause X"} implies \textit{neg=TRUE}) if any;
   \item[2.2.4] verbal form, which can be:
    \begin{itemize}
        \item active (e.g., \textit{X will cause Y});
        \item passive (e.g., \textit{Y will be caused by X});
        \item end-of-sentence (e.g., \textit{Y that  X will cause});
    \end{itemize}
        \end{enumerate}
    \item[2.3] For each sentence with at least one RegEx matching we use the previously extracted information together with a set of verbal form specific functions to split (and reorder) the different components of each sentence in one ore more cause-effect relation triplets, each of which includes the following elements\footnote{The following examples are based on the sentence: \textit{A no-deal Brexit would certainly destroy UK's economy and labour market.}}:
        \begin{itemize}
         \item a subject representing the \textit{cause-side} of the relation (e.g., \textit{A no-deal Brexit ...});
         \item a predicate representing the \textit{relation type} to which the verb phrase corresponds (e.g., \textit{...would certainly destroy...}$\xrightarrow{}$ \textit{rel.type=Destruction});
         \item an object representing the \textit{effect-side} of the relation (e.g., \textit{...UK's economy and labour market.}). 
          \end{itemize}
         \item[2.4] We associate to each cause-effect relation triplet a set of covariates characterising the relation;

    \end{enumerate}
\end{enumerate}
We hence obtain a set of cause-effect relation triplets and their covariates. Each triplet $i$ is characterised by:
\begin{itemize}
    \item $t_i$ - date-time of relation $i$ (which is the publishing date-time UTC of the tweet from where the relation $i$ was extracted)
    \item  $rel.type_i$ - verb phrase type of relation $i$  (3-levels categorical: \textit{Creation},  \textit{Causation}, \textit{Destruction})
    \item $neg_i$ - verb negation dummy of relation $i$ (binary):
    \begin{itemize}
        \item  $neg_{i}=TRUE$ if the verb phrase of the relation $i$ contains a negation;
        \item otherwise $neg_{i}=FALSE$;
    \end{itemize}
    \item $fct_i$ - faction (3-levels categorical: \textit{Brexiteer}, \textit{Remainer}, \textit{Other}):
    \begin{itemize}
        \item $fct_{i}=Brexiteer$ if the biography of the user that has posted the tweet from which the relation $i$ was extracted matches at least one -case insensitive- RegEx condition contained in the \textit{Brexiteer} dictionary ($dict_{Brexiteer}=\{$\textit{brexiteer, vote brexit, voted brexit, voted for brexit, ukip, brexit party, vote leave, leave the EU,respect my vote, johnson, farage, anti-eu,antieu}$\}$) and none of the conditions contained in the \textit{Remainer} dictionary; ($dict_{Remainer}=\{$\textit{remainer, vote remain, voted remain, voted for remain, remain party, new vote, stay in the eu, pro-eu, proeu}$\}$)
        \item  $fct_{i}=Remainer$ if the biography of the user that has posted the tweet from which the relation $i$ was extracted matches at least one -case insensitive- RegEx condition contained in $dict_{Remainer}$ and none of those contained in $dict_{Brexiteer}$;
        \item otherwise  $fct_{i}=Other$.
  %      \item $lnd$ - user declared location is London (binary):
 %   \begin{itemize}
 %       \item  $lnd_{i}=TRUE$ if the declared location of the user that has posted observation $i$ ($loc_i$) contains contains "London";
 %       \item otherwise $lnd_{i}=FALSE$;
 %   \end{itemize}
    \end{itemize}

\end{itemize}

If on one side, the values of the covariates $rel$ and $neg$ are inferred through RegEx functions applied to the Tweet's text (i.e., the post's content), on the other, the faction covariate $fct$ is inferred from the biographical information that a user has written about himself. Hence, in this work, we call \textit{Brexiteers} and \textit{Remainers} those users that self-identify with one of those factions and openly declare it in their Twitter profile Bio. Even though this condition is rather stringent, it allows to minimize the risk of including ``false positives" in our two partisan factions. Besides, being more demanding than other methods based on retweet and following networks, our partisan self-identification method allows for faction changes: a tweet and the cause-effect relations therein contained are considered to have been posted by a partisan user if, at the moment the tweet was posted, the bio of the user posting it matched one of the two partisan faction dictionary conditions. As a result, this framework allows Twitter users that take part to the no-deal debate to dynamically enter and exit a partisan faction, as the Brexit debate and the self-declared faction of users taking part in it change across time.

\subsection{Aggregating arguments and estimating covariate effects}
Through the previous steps, we obtained a set of 204648 relations, each containing a cause-side and an effect-side, hereinafter simply called  \textit{cause} and \textit{effect}. Since in this work we focus on the declared effects of a no-deal scenario, we filter out all extracted relations whose \textit{cause} doesn't match specific RegEx conditions used to verify the presence of \textit{no-deal} in the subject of the relation\footnote{among others, the cause side must match the RegEx "no[- ]?deal", see Table 6 in the Supplement for details.}.
We obtain a set of 36116 relations that match the aforementioned RegEx. 

To aggregate the extracted no-deal effects in classes of effects (i.e., topics) and to see in which terms the propensity to speak about these classes may depend on covariates, we estimate a Structural Topic Model (STM) using only the previously extracted 36116 effects related to the no-deal. 
We selected STM for its unique combination of features required to fulfill our objectives. (i) First, being an extension of  the  Dirichlet-Multinomial Regression topic model\cite{mimno2008topic},STM allows for the inclusion of covariate information in the estimation process. This affects the estimation through informative priors and, more importantly, allows us to evaluate the effects of extracted covariates on topic proportions. (ii) Second, being constructed upon the Correlated Topic Model\cite{blei2006correlated} it allows us to infer the interdependence structure among topics that co-occur in (the effect-side of) relations that have no-deal as a subject. (iii) Finally, being a generalization of the Sparse Additive Generative\cite{eisenstein2011sparse} topic model, it allows covariates to affect the contents of a topic, through sparse deviations with respect to a baseline distribution. 
This feature is here applied to the $fct$ covariate, to model and analyse in which terms, for a given topic, faction-specific argumentative styles can be distinguished from one another.
%by their different usage of words and word associations, that are here modelled through bigrams (i.e., sequences of pairs of words).

Each effect $d\in \{1,...,D\}$ (where $D=36116$) is represented as a set of tokens from a vocabulary of unigrams and bigrams, indexed by $v\in \{1,...,N\}$. 
Effects are hence transformed in matrix called $\mathbf{ExT}$ of size $D-$by$-K$  containing the counts of the number of tokens by effect. 
As a modelling strategy, we allow covariates contained in the $(D)-$by$-(4)$ matrix  $\mathbf{X}=\{t, rel, neg, fct \}$ to affect topic proportions, whereas only the covariate vector $Y=\{fct\}$ is allowed to affect the contents of topics. 
The choice of having the faction covariate $fct$ affect contents is related to our objectives of understanding if and in which terms partisan factions taking part in the Brexit debate use different words (unigrams) and associations (bigrams) to speak about an inferred class of no-deal effects (i.e., topic). 
As tokens, we include unigrams and bigrams which appear at least 10 times in the final collection of no-deal effects. 
This leaves us with a vocabulary $V$ made of $N=3505$ tokens, of which $N_{u}=2462$ are unigrams and $N_{b}=1043$ are bigrams.  \\
The matrices $\mathbf{ExT}$ and $\mathbf{X}$ are hence used, together with the vector $Y$, as inputs to estimate our model using the \textit{Stm} package for R\cite{roberts2019stm}.\\

A Structural Topic Model with $K$ topics is defined as: \\
\begin{center}
\textbf{Topic proportion}
\end{center}
\begin{equation}
\begin{aligned} \mu_{d, k} &=X_{d} \gamma_{k} \\ \gamma_{k} & \sim \mathcal{N}\left(0, \sigma_{k}^{2}\right) \\ \sigma_{k}^{2} & \sim \operatorname{Gamma}\left(s^{\gamma}, r^{\gamma}\right) \end{aligned}
\end{equation}
\vspace{0.7em}
\begin{center}
\textbf{Language model}
\end{center}
\begin{equation}
\begin{aligned}
\theta_{d} \sim LogisticNormal \left(\mu_{d}, \Sigma\right) \\ z_{d, n} & \sim \operatorname{Mult}\left(\theta_{d}\right) \\  v_{d, n} & \sim \operatorname{Mult}\left(\beta_{d}^{k=z_{d, n}}\right) 
\end{aligned}
\end{equation}
\vspace{0.3em}
\begin{center}
\textbf{Topic content}
\end{center}
\begin{equation}
\begin{aligned}
\beta_{d, v}^{k} \propto \exp \left(m_{v}+\kappa_{v}^{, k}+\kappa_{v}^{y,}+\kappa_{v}^{y, k}\right)\\ \kappa_{v}^{y, k} \sim \operatorname{Laplace}\left(0, \tau_{v}^{y, k}\right)\\ \tau_{v}^{y, k} \sim \operatorname{Gamma}\left(s^{\kappa}, r^{\kappa}\right)
\end{aligned}
\end{equation}
Where topics are index by $k$, $X_d$ is a $1-$by$-4$ vector, $\gamma_{k}$ is a $4-$by$-K$ matrix of coefficients, and $\Sigma$ is a $K-$by$-K$ topic proportion covariance matrix.
The distribution over tokens $n$ is the combination of three effects: a topic effect ($\kappa_{v}^{, k}$); a $fct$ covariate effect ($\kappa_{v}^{y,}$), and a topic-covariate interaction effect ($\kappa_{v}^{y,k}$). These three effects are modelled as sparse deviations from a baseline token frequency ($m_{v}$).
To choose the number of topics $K$, we estimate the model for different values of $K$ ranging from $3$ to $70$. For each value of $K$, we repeat 50 times the following procedure: (i) split the the corpus in a random training set and a test set\footnote{the training set contains a random sample containing 25\% of the total number of no-deal \textit{effects}} using a different random seed at each repetition; (ii) estimate the STM model\footnote{See Section 2 in the Supplement for details about STM parameter values.}; (iii)  compute the lower bound and the mean likelihood of the STM to evaluate its performance.
Then for each $K\in\{3, ...,70\}$, the average values of the lower bound and the mean likelihood are computed.  The aforementioned model performance indicators suggest that $K=38$ is a good candidate number of topics for estimating an STM with our corpus of no-deal \textit{effects}\footnote{see Section 2 in the Supplement.}. 
Finally, the STM is re-estimated for $K=38$ with the whole set of \textit{effects}\footnote{Using spectral initialization, which allows the estimated STM to be deterministic conditionally on parameters and covariates values. }.

% topic contents (distribution of unigrams and bigrams in each topic)
%To determine the number of topics we use the 

\subsection{Constructing faction-specific narrative networks}\label{sec:method:intra_topic}
To construct faction specific narrative networks for a topic $k$ we use the posteriors of $m_{v}$, $\kappa_{v}^{, k}$, $\kappa_{v}^{y,}$, $\kappa_{v}^{y, k}$, which are respectively called  $\hat{m}_{v}$, $\hat{\kappa_{v}}^{, k}$, $\hat{\kappa_{v}}^{y,}$, $\hat{\kappa_{v}}^{y, k}$. In particular, we first separate tokens in the vocabulary $V$ in two disjoint sets $V_u$ and $V_b$, where $V_u$ contains only the unigrams from the vocabulary $V$ and $V_b$ only the bigrams. For all token $v\in V_u$, we filter out the less relevant unigrams  for the $k$th topic and for faction \textit{Others}. We do so by keeping only tokens above the 80\textit{th} percentile rank, in terms of the following posteriors sum: $\hat{m}_{v}+ \hat{\kappa_{v}}^{, k}+\hat{\kappa}_{v}^{y=Others,}+\hat{\kappa}_{v}^{y=Others, k}$.  We then apply the same procedure for bigrams (all $v\in V_b$). Finally we filter out bigrams that are not connecting unigrams pairs in the 20\% top percentile. From the resulting unigram and bigram sets and their weights, which are given by the exponential of the aforementioned posteriors sum, we can construct the narrative network of the topic $k$ for non partisan users ($fct=Others$). We construct it by using unigrams as nodes and bigrams as edges, and by representing unigrams' weights through the node size and bigrams' weights through the edge width. The resulting network can be seen as a graphical representation of the phrasal microstructure of the debate about a topic, for non partisan users. Since we are interested in analysing how partisan factions intervene in this debate we overlay the topic content differences between the two partisan factions using a continuous color scale ranging from blue (for negative values) to light-gray (for zero) to red (for positive values). For any token $v$, the posterior probability difference between faction $i$ and faction $j$ is given by:\\
\begin{equation}
    \hat{\delta}_{v, i,j}= exp(\hat{m}_{v}+\hat{\kappa_{v}}^{y=i,}+\hat{\kappa_{v}}^{y=i, k})-exp(\hat{m}_{v}-\hat{\kappa_{v}}^{y=j,}+\hat{\kappa_{v}}^{y=j, k})
\end{equation}
With $i=Brex.$ and $j=Rem.$, we obtain the difference between the \textit{Brexiteers} and Remaines partisan factions for the topic $k$. We can hence overlay the colors representing partisan factions' differences to the narrative network of non-partisan users (i.e., \textit{Others}), as shown in Figure \ref{fig:topiczoom} for topic 2.

\subsection{Filtering the network structure of a debate and identifying constellations of effect-classes}\label{sec:method:aggregatedebate}

To explore the relationship between covariate values and topic proportions  we use the \textit{estimateEffect} function of the STM library. This function allows to estimate the effects of one or more covariates included in the STM estimation phase on expected topic proportions. For each \textit{effect} $d$, the proportions of a topic $k$ are modelled as a function of the faction ($fct_d$), relation type ($rel.type_d$), and verb negation ($neg_d$) covariate values:
\begin{equation}
\begin{aligned}
    {propensity}_{k,d}=f({fct}_d,{neg}_d,{rel.type}_d)
%= X_d \gamma_k \\
%    =Const_{k}+\gamma_{k,fct} fct_d +\gamma_{k,neg}neg_d+\gamma_{k,rel.type}rel.type_d
    \end{aligned}
\end{equation}
This method also allows to asses which covariate coefficients are statistically significant\footnote{see Table 3 in the Supplement for regression results}.
%This method also allows us also to asses for which covariates and covariate values\footnote{we recall that for each non-binary covariate $i$, e.g., $fct$  and $rel.type$, $\gamma_{k,i}$ is a vector of $n-1$ coefficients, where $n$ represents the number of levels the categorical variable $i$ can take and $i_d$ is a vector of binary variables (switches) representing which non-deafault level of the covariate $i$ is matched . The default covariate level for $fct$ is $Brex.$, and for $rel.type$ is $Causation$}  the estimated coefficients are statistically significant\footnote{see table {\color{red}[reference to table]} in the Supplement for regression results}.
To analyse if there are significant differences in topic prevalence among the two partisan factions (\textit{Brexiteers} Vs \textit{Remainers}) and among the two polarized verb types (\textit{Creation} Vs \textit{Destruction}), we compute the coefficients' differences and their variance to test if the former are statistically different. 

To represent the aggregate structure of the debate about no-deal Brexit effects, we transform the $\hat{\Sigma}$ matrix in a topics propensity correlation matrix.
We hence obtain a pruned correlation matrix that can be visualized as an undirected graph, where nodes represent topics and edges represent correlations between them. We then label the correlation graph on the basis of covariate-levels that have predominant effects (i.e., significantly larger coefficients) with respect to their opposing type (\textit{Brex} Vs \textit{Rem.}, \textit{Creation} Vs \textit{Destruction}). More specifically, for each topic we label its node on the basis of the covariate level that implies a significantly higher propensity for that topic (if any), otherwise we label that predominance relation property with the ``none" label.
The topic's predominant relation type ($\hat{\beta}_{rel.type=Destruction}$ Vs $\hat{\beta}_{rel.type=Creation}$) is represented through the node color. Whereas, the topic's predominant partisan faction ($\hat{\beta}_{rel.type=Brex.}$ Vs $\hat{\beta}_{rel.type=Rem.}$) is represented through the node shape. Finally, we represent significant positive effects on a topic's propensity related to the presence of a negated relation (i.e., $\hat{\beta}_{neg=TRUE}>0$) by applying a shadow around the topic node.  
To highlight the main structural relations among topics used in the no-deal effects debate, and identify topic constellations that attract or repulse each other, we filter the correlation graph using different threshold levels applied to the absolute value of the correlations, which are represented through the edges' width. These threshold values are progressively lowered, and at each step isolated nodes are removed to show only the backbone of the debate for that specific filtration level. This allows to analyse the building blocks of a debate and how these blocks grow as we lower the threshold. Moreover, by analysing the topology of this network (e.g., balanced and unbalanced triangles or cliques) one can see if these building blocks are coherent either in terms of the sign of the correlations that characterise them, or in terms of the property labels associated to predominant covariate levels.

\subsection{Identifying time-varying faction lead/follow relations}\label{sec:method:lead-follow}

Many methods to analyse lead-follow relations among time series exist, like cross-correlations among faction-specific daily topic proportions series. Despite their usefulness, these methods have several limits, in particular the resulting lead-follow relations are by construction static. As a result, we apply a method based on Dynamic Time Warping\cite{berndt1994using}(DTM), which allows inferred lead-follow relations among factions to change across time.

As a first step to identify the time varying faction lead/follow relations, the  $D-$by$-K$ matrix containing the distribution of topics by effect, called $\mathbf{ExT}$, is extracted from the estimated STM. We have that $\mathbf{ExT}_{d,k}$ represents the estimated propensity of topic $k$ in the no-deal \textit{effect} $d$, and $\sum_{j=1}^{K} \mathbf{ExT}_{d,j}=1$.
Using $\mathbf{ExT}$ together with the $t$ and $fct$ covariates contained in $\mathbf{X}$, for each faction $i\in\{Brex., ..., Rem.\}$, for each topic $k\in\{1, ..,K\}$, and for each day $x\in\{01-02-2019,02-02-2019,...,01-05-2020\}$, we compute the average daily propensity of topic $k$ for faction $i$ on day $x$ and call it $w_{x,k,i}$. 
Where for a specific day $x$ and faction $i$, we have that $w_{x,k,i}>0\; \Finv k$ and $\sum_{j}w_{x,j,i}=1$ if there is at least one effect \textit{d} that has \textit{x} as date ($t_d$ covariate equal to $x$) and \textit{i} as faction ($fct_d$ covariate equal to $i$) \footnote{i.e., $card(t==x \otimes fct==i)>0$}, and $w_{x,k,i}=0 \; \forall k$ otherwise\footnote{$w_{x,k,i}=0 \; \forall k \; \Rightarrow\sum_{j}w_{x,j,i}=0$}.

Each matrix $w_{.,.,i}$ is of size $T-$by$-K$ and contains as column vectors $K$ time-series with the average estimated topic propensities of extracted effects posted by users belonging to faction $i$. 
Dynamic lead-follow relations among factions are identified with a DTM method called FLICA\cite{amornbunchornvej2018framework}. In this framework the notion of leading entity (i.e., leading faction) corresponds to the initiation of topical proportion patterns that other factions hence follow. Given a set of time series representing average topic proportions for each faction, one can use this method to identify periods of coordinated activity between factions, and infer the dynamics  across time of lead and follow relations between them. The algorithm takes as input the $w_{.,.,i}$ matrices for two or more factions, for example  \textit{Brexiteers} ($i=Brex.$) and \textit{Remainers} ($i=Rem.$), each of which can be seen as a 38-dimensional time series (at the daily frequency), and through a DTM algorithm gives as output a dynamic directed network, for which nodes represent the factions and edges represent following relations between them. Each frame of this dynamic network represents a day. For each frame, inferred lead-follow relations between pairs of nodes are mutually exclusive, so, or \textit{Brexiteers} follow \textit{Remainers} ($Brex. \rightarrow Rem.$ and $Brex. \nleftarrow Rem.$) or \textit{Remainers} follow \textit{Brexiteers} ($Brex. \nrightarrow Rem.$ and $Brex. \leftarrow Rem.$)  or no lead-follow relation is observed ($Brex. \nrightarrow Rem.$ and $Brex. \nleftarrow Rem.$) .  $f_{i,j,x}\in]0,1]$ is the weight of the edge $ij$ at the date $x$, and represents the strength of the follow relation (if any) between node $i$ and $j$ at a specific day. 
The values of $f_{Brex.,Rem.,x}$ and $f_{Rem., Brex.,x}$ are used to build our partisan faction \textit{lead/follow relation index} (central plot in Figure \ref{fig:lead_follow}), which is defined as follows:
\begin{equation}
fl_{Rem.,Brex.,x}=f_{Rem., Brex.,x}-f_{Brex.,Rem.,x}\in[-1,1]
\end{equation}
To implement the FLICA algorithm, we employ mFLICA function from the mFLICA library for R\cite{amornbunchornvej2020mflica}, using a \textit{window} of one month (30 days), a \textit{max lag window} of 6 days, and a \textit{window time shift} of 1 day. Results appear to be robust, and small and medium changes in the aforementioned parameters give similar results.

\section*{Acknowledgements}
%Acknowledgements should be brief, and should not include thanks to anonymous referees and editors, or effusive comments. Grant or contribution numbers may be acknowledged.

The authors acknowledge financial support from the European Union ODYCCEUS Horizon 2020 project, grant agreement number 732942.

\section*{Author contributions statement}
M.W.  and C.S.  conceived the paper, the methodology,  analysed the results and wrote the paper.  C.S. downloaded the data and undertook the data analysis. M.W.  and C.S.  reviewed the manuscript. 
\onecolumngrid
\appendix
\section*{Additional information}

\begin{table}[!htbp]
\caption{\label{tab:relationcounts}%
Counts (and shares) of extracted relations that have "no deal" in cause-side, by relation type and by faction, followed by Pearson's Chi-squared test
}
{\hspace*{-0.5cm}
\begin{tabular}{|r|ccc|c|}
\hline
  & Creation & Causation & Destruction & \textbf{TOT.} \\
\colrule
Brexiteers & 436 (49.7\%) & 325 (37.1\%) & 116 (13.2\%)  & 877 (2.4\%)  \\\cline{2-4}
Others & 15817 (46.0\%) & 13181  (38.3\%) & 5398 (15.7\%)  & 34396 (95.2\%)\\\cline{2-4}
Remainers & 361 (42.8\%)  & 334  (39.6\%)  & 148 (17.6\%)  & 843 (2.3\%) \\\cline{2-4}
\colrule
\textbf{TOT.} & 16614 (46.0\%) & 13840  (38.3\%) & 5662 (15.7\%) & 36116 \\\hline
\end{tabular}}\newline
Pearson's Chi-squared test (vars: \textit{faction} and \textit{rel. type}, data: \textit{causal relations that "no deal" in cause-side}):\newline
X-squared $= 10.482$, df $= 4$, p-value $= 0.03305$
\end{table}

 \begin{table}[!htbp]
\caption{\label{tab:relationcounts}%
Count (and share) of tweets and retweets about the ``no-deal''  containing one or more \textit{Causal Markers} (CM), by faction, followed by Pearson's Chi-squared tests.}
\begin{tabular}{|r|c|c|}
\hline
  &  TWEETS with CM$^{1}$ & TWEETS without CM$^{1}$ \\
\colrule
Brexiteers & 7407 (11.9\%)  & 54900 (88.1\%)  \\
Others &  265879 (12.5\%)   & 1853757 (87.5\%)  \\
Remainers & 6304 (15.2\%)  & 35161 (84.8\%) \\\hline
\end{tabular}\newline
Pearson's Chi-squared test (vars: \textit{faction} and \textit{contains.causal.marker}, data:\textit{ tweets only}):\\
X-squared $= 289.18$, df $= 2$, p-value $< 10^{-15}$
\vspace{1em}\newline
\begin{tabular}{|r|c|c|}
\hline
  &  RETWEETS with CM$^{1}$  & RETWEETS without CM$^{1}$ \\
\colrule
Brexiteers & 28928 (14.5\%) &  170617 (85.5\%) \\
Others &  1024830 (16.0\%)  &  5394586  (84.0\%) \\
Remainers &  27630 (17.0\%) & 134928 (83.0\%)  \\\hline
\end{tabular}
\newline
Pearson's Chi-squared test (vars: \textit{faction} and \textit{contains.causal.marker}, data:\textit{ retweets only}):\\
X-squared $= 448.21$, df $= 2$, p-value $< 10^{-15}$
\vspace{1em}\newline
%\textbf{ $^{1}$ The list of \textit{Causal Markers} used to construct the\textit{ contains.causal.marker} dummy variable is presented in section {\color{red}[add reference to methodological section on causal markers]}}
\end{table}

\begin{table}[h]
\caption{\label{tab:counts}%
Counts (and shares) of collected Twitter posts about "no deal" by type and by faction
}
\begin{tabular}{|r|cc|c|}
  \hline
  & TWEETS & RETWEETS & \textbf{FACTION TOT.} \\
\colrule
Brexiteers & 62307 (23.8\%) & 199545 (76.2\%)  & 261852 (2.9\%)\\ \cline{2-3}
Others & 2119636 (24.8\%) & 6419416 (75.2\%) & 8539052 (94.8\%) \\\cline{2-3}
Remainers & 41465  (20.3\%)  & 162558  (79.7\%)  & 204023 (2.3\%) \\\cline{1-4}
\textbf{TYPE TOT.}  & 2223408 (24.7\%) & 6781519 (75.3\%) & 9004927 \\\hline
\end{tabular}\\
Pearson's Chi-squared test (vars: \textit{faction} and \textit{is.retweet}):\\
X-squared $= 2285.7$, df $= 2$, p-value $< 10^{-15}$
\end{table}

\begin{table}[!htbp]
\caption{\label{tab:relationcounts_appendix}%
Counts of extracted relations that have "no deal" as cause, by relation type, negation dummy and faction, followed by Mantel-Haenszel chi-squared tests
}
\centering
\begin{tabular}{|lclcrr|}
  \hline
 \textit{fct} & N. (\%) &  \textit{negated} & N. (\%) &\textit{ rel.type}  & N. (\%) \\ 
  \hline
\multirow{6}{2.75cm}{Brexiteers} & \multirow{6}{2.5cm}{877 (2.4\%)}  & \multirow{3}{2.5cm}{FALSE} &  \multirow{3}{2.5cm}{856 (97.6\%)} & Creation &  423 (49.4\%)\\
 & & &  & Causation & 319 (37.3\%)\\ 
 & & &  & Destruction &  114 (13.3\%)\\\cline{5-6}
 &  &  \multirow{3}{2.5cm}{TRUE} & \multirow{3}{2.5cm}{21 (2.4\%)}  & Creation &   13 (61.9\%) \\ 
 & & &  & Causation &   6 (28.6\%) \\ 
 & & &  & Destruction &    2 (9.5\%) \\\cline{3-6}
\multirow{6}{2.75cm}{Others} & \multirow{6}{2.5cm}{34396 (95.2\%)}  & \multirow{3}{2.5cm}{FALSE} & \multirow{3}{2.5cm}{33343 (96.9\%)} & Creation &  15162 (45.5\%) \\ 
 & & &  & Causation &  12892 (38.7\%)  \\ 
 & & &  & Destruction & 5289 (15.8\%)  \\\cline{5-6}  
 &  &  \multirow{3}{2.5cm}{TRUE} & \multirow{3}{2.5cm}{1.053 (3.1\%)}  & Creation &  655 (62.2\%)\\ 
 & & &  & Causation &  289 (27.4\%)\\ 
 & & &  & Destruction &  109  (10.4\%) \\ \cline{3-6}
\multirow{6}{2.75cm}{Remainers}  & \multirow{6}{2.5cm}{843 (2.3\%)}  & \multirow{3}{2.5cm}{FALSE} &\multirow{3}{2.5cm}{813 (96.4\%)} & Creation & 344 (42.3\%)\\ 
 & & &  & Causation & 326 (40.1\%) \\ 
 & & &  & Destruction & 143 (17.6\%) \\ \cline{5-6}
 &  &  \multirow{3}{2.5cm}{TRUE} & \multirow{3}{2.5cm}{30 (3.6\%)}  & Creation &  17 (56.7\%)\\ 
 & & &  & Causation &   8 (26.7\%)\\ 
 & & &  & Destruction &   5 (16.6\%) \\ 
  
   \hline
\end{tabular}\\
Mantel-Haenszel chi-squared tests:\\
-  strata: \textit{fct}, vars: \textit{rel.type} and \textit{negated}, $M^2 = 118.81$, df $= 2$, p-value $< 10^{-14}$\\
- strata: \textit{rel.type}, vars: \textit{negated} and \textit{fct}, $M^2 = 2.5141$, df $= 2$, p-value $= 0.2845$\\
- strata: \textit{negated}, vars: \textit{rel.type} and \textit{fact}, $M^2 = 10.994$, df $= 4$, p-value$ = 0.02663$

\end{table}

\end{document}

\title[Supplement - Statistical Rhetoric of Online Arguments \& Debates]{\textbf{Supplement to:}\newline\newline ``\textit{Statistical Rhetoric of Online Arguments \& Debates}''\newline\newline (Dated:  \today)}
\author[C.R.M.A. Santagiustina]{Carlo R. M. A. Santagiustina}
 \address{\color{black}Ca' Foscari University of Venice, Cannaregio 873, 30123 Venice, Italy}
  \email{carlo.santagiustina@unive.it}
\author[M. Warglien]{Massimo Warglien}
\address{\color{black}Ca' Foscari University of Venice, Cannaregio 873, 30123 Venice, Italy}%
 \email{warglien@unive.it}

\renewcommand{\abstractname}{Supplementary material}

%\maketitle

\maketitle
\section*{Supplement content summary}
\begin{figure}
 "CREATION" ARGUMENT(S) \hspace{14em}  "DESTRUCTION" ARGUMENT(S)
\centering{
\includegraphics[trim= 7mm 10mm 0mm 20mm,clip, width= \textwidth]{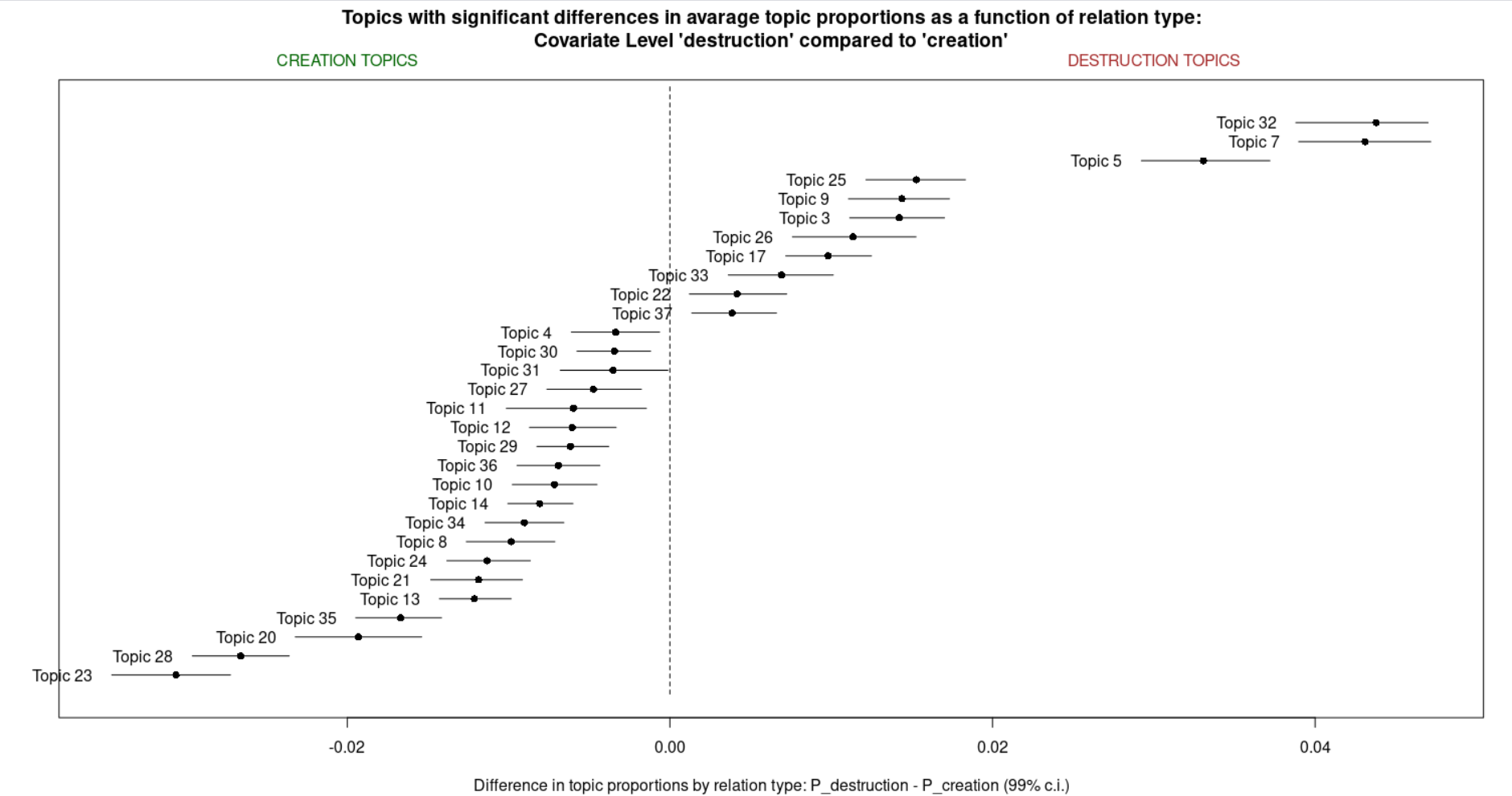}
\caption{Significant estimated differences in argument $k$ proportions (with 99\% C.I.) of as a function of causal relation type $rel$: $\hat{\beta}_{k,rel=Destruction}-\hat{\beta}_{k,rel=Creation}$}.
\label{S_fig:reltype_effect}}
\end{figure}

\begin{figure}
 "REMAINER" ARGUMENT(S) \hspace{14em}  "BREXITEER" ARGUMENT(S)
    \centering{
\includegraphics[trim= 0mm 12mm 0mm 20mm,clip, width= \textwidth]{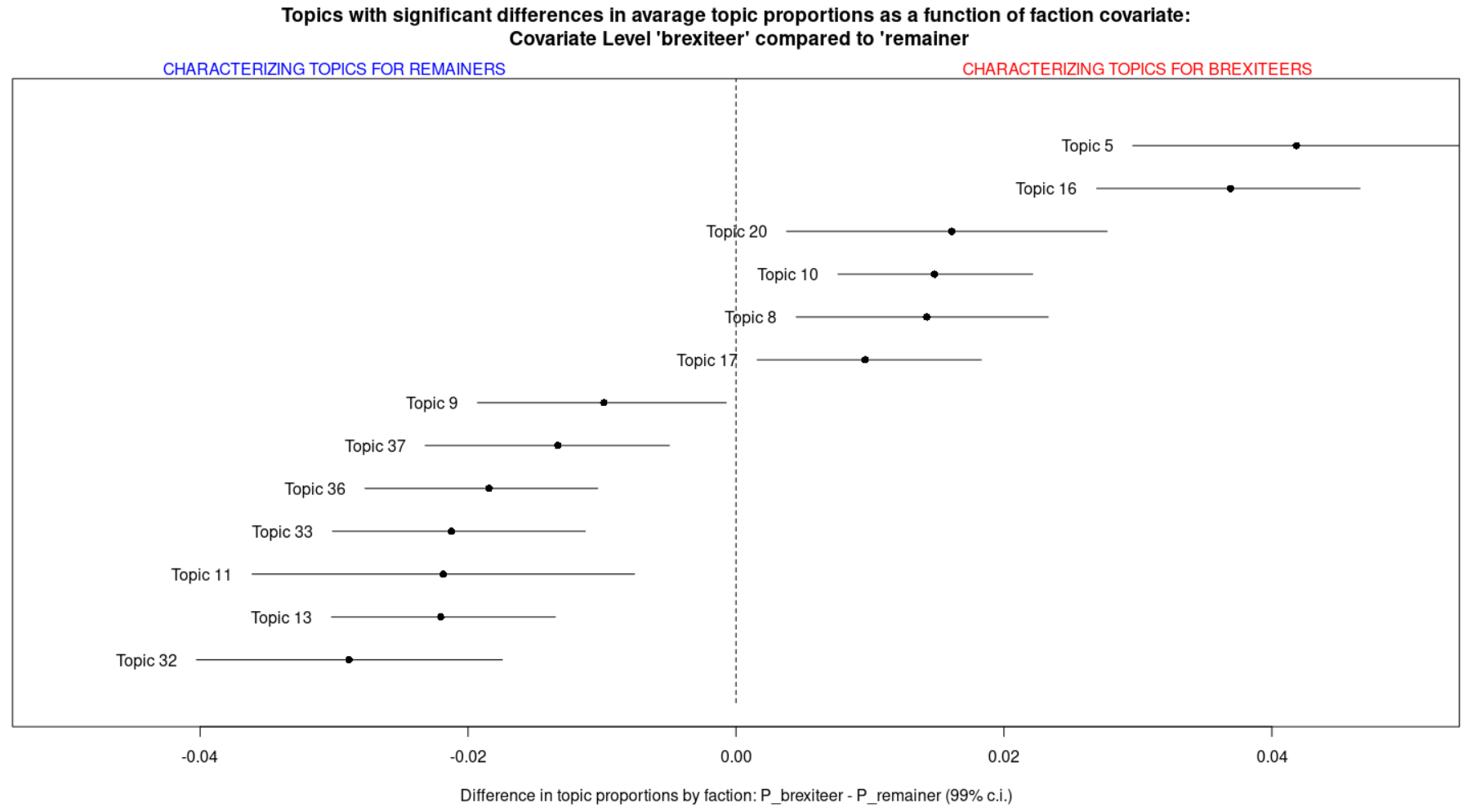} 
\caption{Significant estimated differences in topic $k$ proportions (with 99\% C.I.) of as a function of partisan faction $fct$: $\hat{\beta}_{k,fct=Brexiteer}-\hat{\beta}_{k,fct=Remaniner}$}.
\label{S_fig:faction_effect}}
\end{figure}

\begin{figure}
 NON-NEGATED ARGUMENT(S) \hspace{14em}  NEGATED ARGUMENT(S)
    \centering{
\includegraphics[trim= 0mm 12mm 0mm 19mm,clip, width= \textwidth]{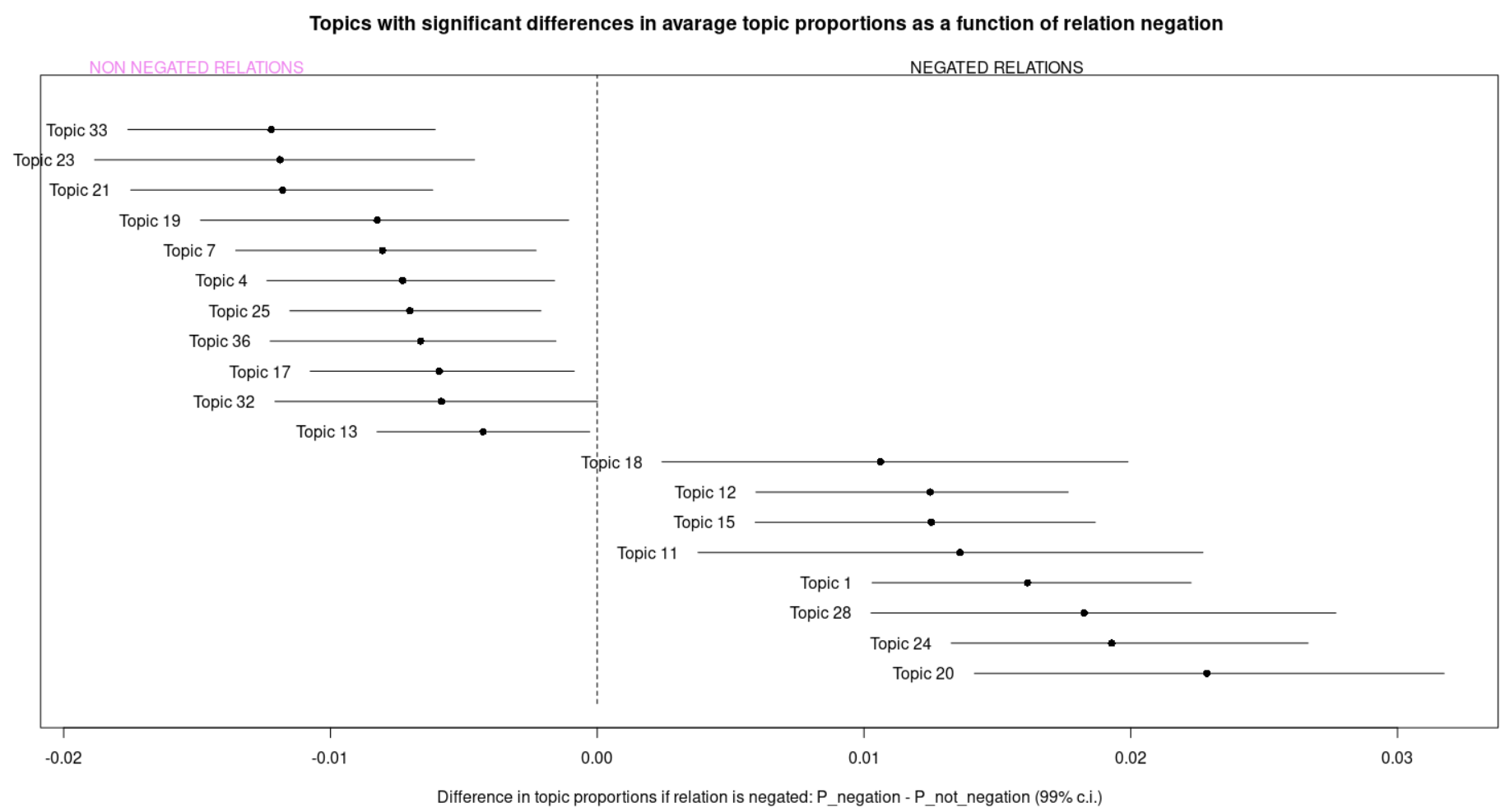} 
\caption{Significant estimated differences in topic $k$ proportions (with 99\% C.I.) of as a function of relation verb negation $neg$: $\hat{\beta}_{k,neg=TRUE}-\hat{\beta}_{k,neg=FALSE}$}.
\label{S_fig:negation_effect}}
\end{figure}

\begin{figure}
\includegraphics[trim= 0mm 0mm 0mm 0mm,clip, width=1\textwidth]{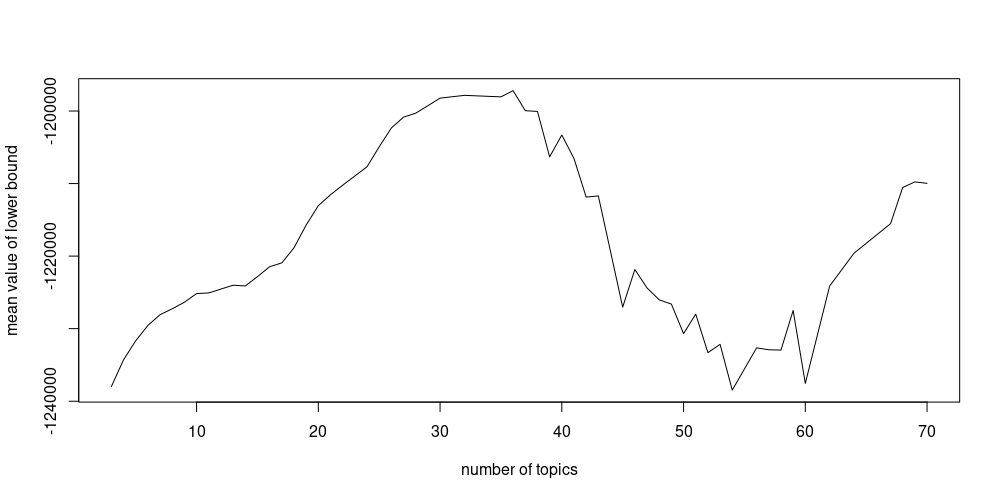}
\caption{Mean values of the STM \textit{Lower bound} as a function of $K$. At each replication a different random sample containing a 25\%  of the population of no-deal effects has been heldout.}
\label{S_fig:heldout_lb}
\end{figure}

\begin{figure}
\includegraphics[trim= 0mm 0mm 0mm 0mm,clip, width=1\textwidth]{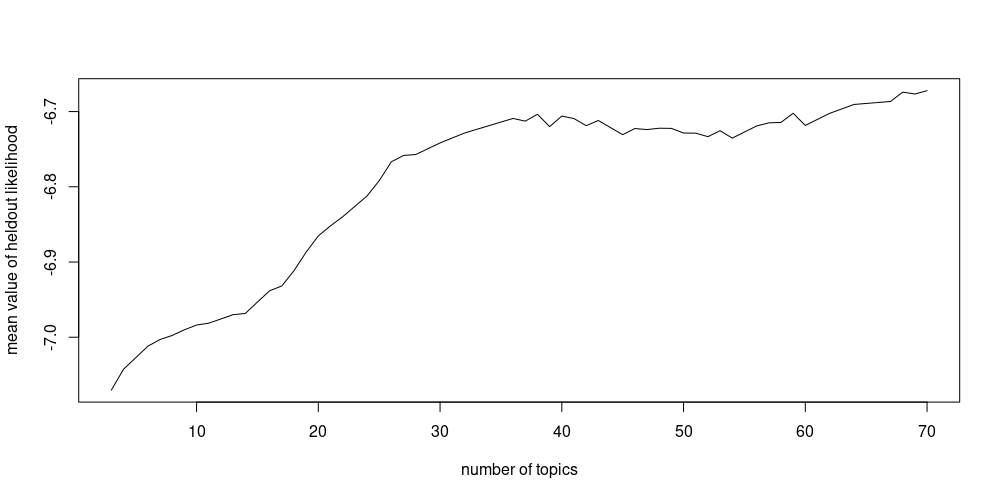}
\caption{Mean values of the STM \textit{Heldout likelihood}  as a function of $K$. At each replication a different random sample containing a 25\%  of the population of no-deal effects has been heldout. The heldout sample was hence used to compute the \textit{Heldout likelihood}.}
\label{S_fig:lb_likelihood}
\end{figure}

\begin{figure}
\includegraphics[trim= 0mm 0mm 0mm 0mm,clip, width=1\textwidth]{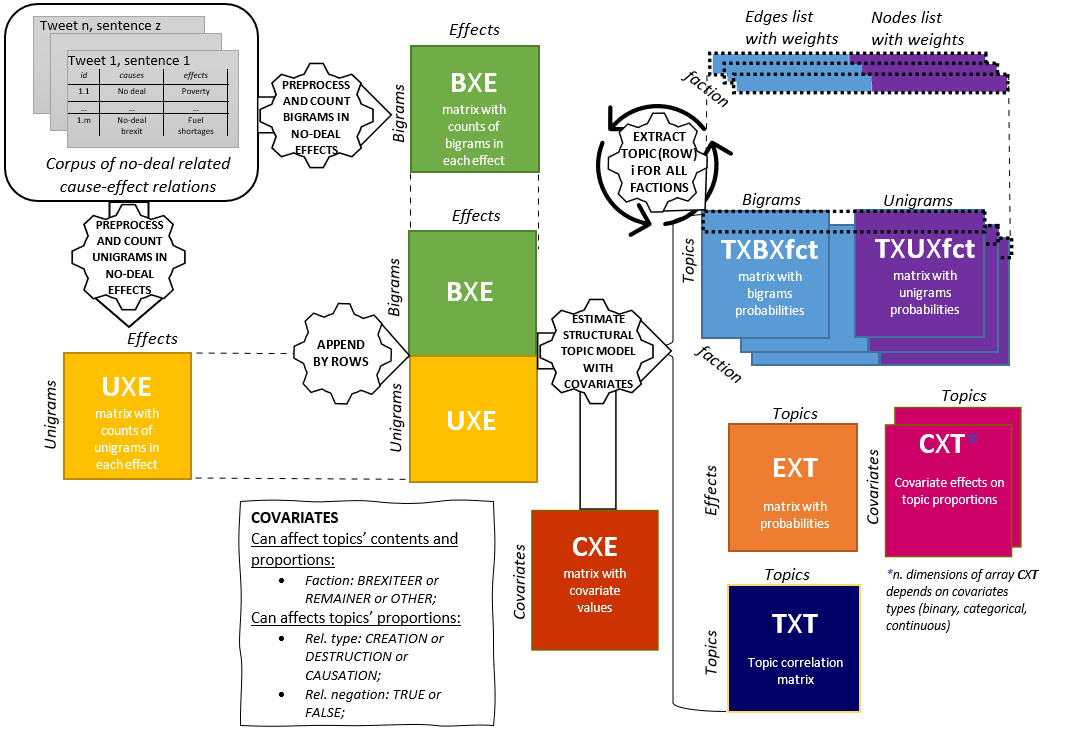}
\caption{Workflow summary}
\label{S_fig:workflow}
\end{figure}

Supplement Section~\ref{S_sec:Results} contains additional materials related to the results.  It includes: the values of the parameters used for the final STM estimation (Table \ref{S_tab:STM_parameters}); a summary table of top 10 tokens by topic and by faction (Table \ref{S_tab:summary_table}); a table containing estimated topic propensity covariate effects (Table \ref{S_tab:topic_prop_est}); three figures representing respectively significant differences in topic proportions as a function of polar verb type (Figure \ref{S_fig:reltype_effect}), partisan faction (Figure \ref{S_fig:faction_effect}), and negation (Figure \ref{S_fig:negation_effect}) covariates.\\
Robustness checks and description of parameters used in the STM are described in Section~\ref{S_sec:Robustness} of the Supplement.  The section includes: the values of the parameters used in experiments for choosing the number of topics $K$ (Table \ref{S_tab:K_choice_parameters}); two Figures representing the STM's \textit{Lower bound} (Figure \ref{S_fig:lb_likelihood}) and \textit{Heldout likelihood} (Figure \ref{S_fig:heldout_lb}) as a function of $K$.\\
Finally,  details about the methods and code are given in Section~\ref{S_sec:Methodology}, which contains: a summary of the workflow (Figure \ref{S_fig:workflow}); the list of used stopwords (Table \ref{S_tab:stopwords}); a RegEx used to identify relations that mention the no-deal in the cause-side (Table \ref{S_tab:nodeal_regex}); the list of verbs used to build verb phrases related to \textit{Destruction}, \textit{Causation} and \textit{Creation} (Table \ref{S_tab:verbslist}); and the R code used to identify extract cause-effect relations from texts (Table \ref{S_tab:rel_extraction_RegEx}).\\

\section{Results supplement}\label{S_sec:Results}

The results in the main paper and in the Supplement have been obtained using the parameter values in Table \ref{S_tab:STM_parameters} for estimating the STM. For not listed parameters, \textit{Stm} (V.1.3.5) library defaults have been used. The model converged after 32 iterations. The value of the \textit{Lower Bound} at the final iteration is $-1203368.488309473032132$.
\begin{table}[]
    \centering
        \caption{Parameter values used in the final STM estimation}
    \label{S_tab:STM_parameters}
    \begin{tabular}{|c|c|}
    \hline
    \textbf{parameter} & \textbf{value}\\\hline
       K  & 38\\
       Topic prevalence &  $~ rel.type + neg + fct + s(t)$\\
       Topic content & ~ $fct$\\
         max.em.its & 500 \\
  emtol & $1e-05$ \\
  init.type & Spectral \\
    gamma.maxits & 5000 \\\hline
    \end{tabular}

 %Lower bounds values (by iteration) -1335345.448282827157527 -1299035.588527687126771 -1260271.188331603072584 -1241371.964036633260548 -1230574.514381499262527 -1224295.156499120872468  -1219612.646936177741736 -1216360.528544795000926 -1213908.808361426228657 -1211880.197008469142020 -1210290.567023652140051 -1208997.593114702962339 -1207898.649767846101895 -1207063.861249184468761 -1206368.476130713010207 -1205850.786755161127076 -1205408.155984744429588 -1205053.411238851258531 -1204701.268799427198246 -1204365.264400240965188 -1204137.116125457920134 -1203980.841004661982879 -1203860.253143168753013 -1203776.553669569082558 -1203689.155620086006820 -1203646.259989667916670 -1203584.424415680347010 -1203521.637313834857196 -1203451.766061201225966 -1203406.479604971595109 -1203378.834222923032939 -1203368.488309473032132

\end{table}
\pagebreak
\begin{table}
    \caption{\label{S_tab:summary_table}%
Summary table of topics  (by overall topic proportion), with top 10 tokens (by token probability) by topic and by faction. Row color scale represents significant (at the 0.01 significance level) differences of the estimated topic proportion coefficients of the two partisan factions. The closer is the row color to the corresponding faction color, the more characterizing is a specific topic for one of the two partisan faction.  Characterizing topics for \textit{Remainers} are in {\color{blue} blue}, whereas characterizing topics for \textit{Brexiteers} are in {\color{red} red}.
}\end{table}

\includegraphics[trim= 0mm 0mm 0mm 0mm,clip, width= 1\textwidth]{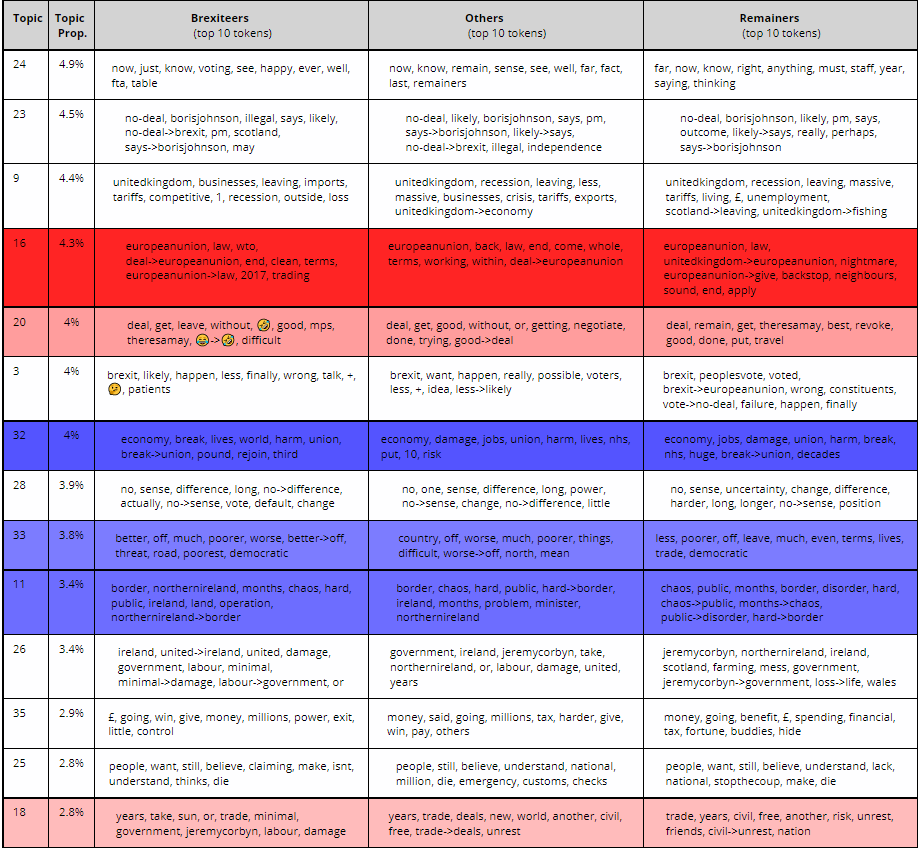}\newline
\begin{center}
    Table 1 --  \textit{Continued on next page}
\end{center}

\newpage
\begin{center}
{ Table 1 -- \textit{Continued from previous page}}
\end{center}
\includegraphics[trim= 0mm 0.5mm 0mm 0mm,clip, width= 1\textwidth]{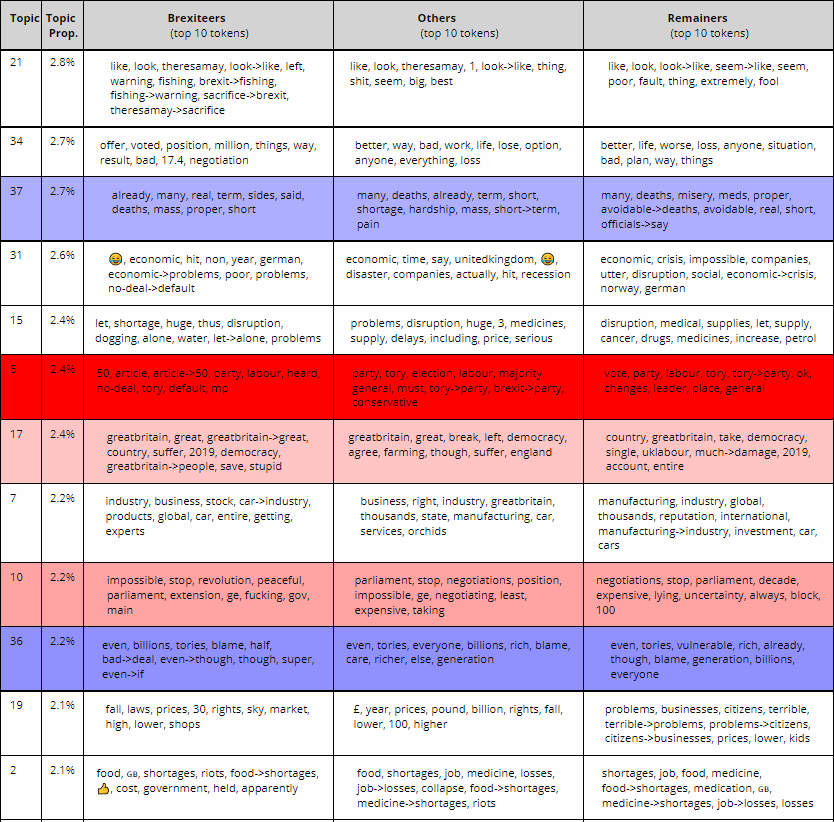}\newline
\begin{center}
    Table 1 --  \textit{Continued on next page}
\end{center}

\newpage
\begin{center}
{ Table 1 -- \textit{Continued from previous page}}
\end{center}
\includegraphics[trim= 0mm 0mm 0mm 0mm,clip, width= 1\textwidth]{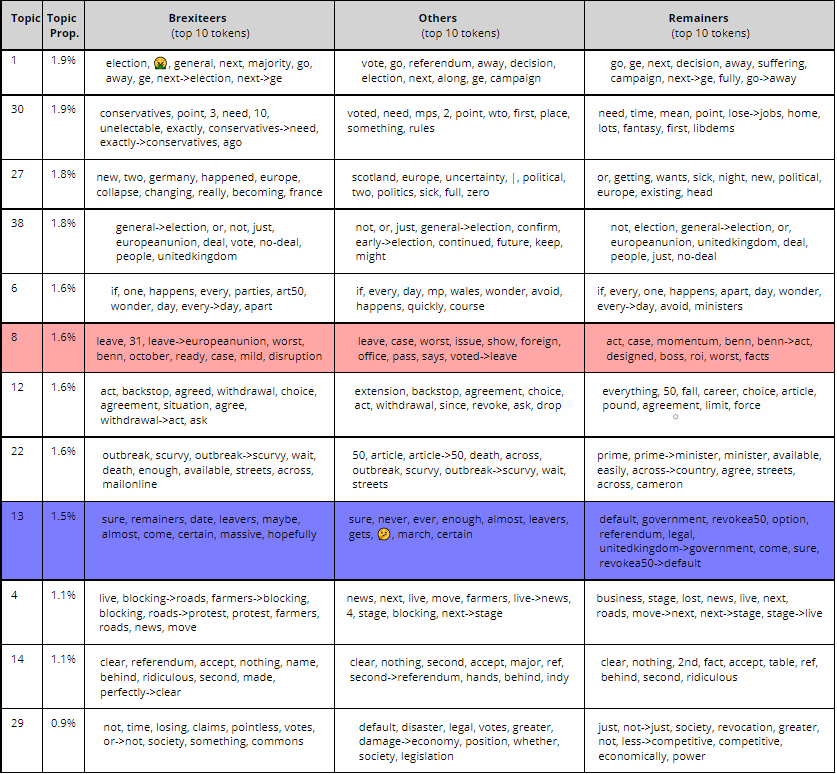}

\newpage
\begin{longtable}{rrlrrrr}
\caption{Estimated topic propensity parameters}\label{S_tab:topic_prop_est}\\
  \hline
 & \textbf{topic} & \textbf{term} & \textbf{estimate} & \textbf{std.error} & \textbf{statistic} & \textbf{p.value} \\ 
  \hline
\endfirsthead
\multicolumn{7}{c}%
{\tablename\ \thetable\ -- \textit{Continued from previous page}} \\
\hline
 & \textbf{topic} & \textbf{term} & \textbf{estimate} & \textbf{std.error} & \textbf{statistic} & \textbf{p.value} \\ \hline
\endhead
\hline \multicolumn{7}{c}{\textit{Continued on next page}} \\
\endfoot
\hline
\endlastfoot
 &     1 & Constant  & 0.0160 & 0.0023 & 7.0273 & 0.0000 \\ 
   &     1 & \textit{rel.type}=Creation & 0.0083 & 0.0008 & 10.8517 & 0.0000 \\ 
   &     1 & \textit{rel.type}=Destruction & 0.0078 & 0.0011 & 7.0078 & 0.0000 \\ 
   &     1 & \textit{neg}=TRUE & 0.0165 & 0.0025 & 6.4734 & 0.0000 \\ 
   &     1 & \textit{fct}=Other & 0.0006 & 0.0023 & 0.2592 & 0.7955 \\ 
   &     1 & \textit{fct}=Remainer & 0.0012 & 0.0034 & 0.3577 & 0.7206 \\ \hline
   &     2 & Constant  & 0.0405 & 0.0032 & 12.5478 & 0.0000 \\ 
   &     2 & \textit{rel.type}=Creation & -0.0394 & 0.0013 & -31.0073 & 0.0000 \\ 
   &     2 & \textit{rel.type}=Destruction & -0.0387 & 0.0016 & -24.5399 & 0.0000 \\ 
   &     2 & \textit{neg}=TRUE & -0.0038 & 0.0028 & -1.3641 & 0.1725 \\ 
   &     2 & \textit{fct}=Other & 0.0088 & 0.0029 & 2.9790 & 0.0029 \\ 
   &     2 & \textit{fct}=Remainer & 0.0062 & 0.0045 & 1.3652 & 0.1722 \\ \hline
   &     3 & Constant  & 0.0221 & 0.0023 & 9.7882 & 0.0000 \\ 
   &     3 & \textit{rel.type}=Creation & 0.0100 & 0.0008 & 12.4265 & 0.0000 \\ 
   &     3 & \textit{rel.type}=Destruction & 0.0244 & 0.0011 & 22.7367 & 0.0000 \\ 
   &     3 & \textit{neg}=TRUE & 0.0042 & 0.0021 & 2.0082 & 0.0446 \\ 
   &     3 & \textit{fct}=Other & 0.0020 & 0.0023 & 0.8617 & 0.3889 \\ 
   &     3 & \textit{fct}=Remainer & -0.0054 & 0.0032 & -1.7141 & 0.0865 \\ \hline
   &     4 & Constant  & 0.0157 & 0.0028 & 5.6770 & 0.0000 \\ 
   &     4 & \textit{rel.type}=Creation & 0.0016 & 0.0009 & 1.7799 & 0.0751 \\ 
   &     4 & \textit{rel.type}=Destruction & -0.0019 & 0.0013 & -1.4109 & 0.1583 \\ 
  &     4 & \textit{neg}=TRUE & -0.0073 & 0.0021 & -3.4211 & 0.0006 \\ 
  &     4 & \textit{fct}=Other & 0.0020 & 0.0029 & 0.6942 & 0.4875 \\ 
  &     4 & \textit{fct}=Remainer & -0.0028 & 0.0038 & -0.7466 & 0.4553 \\ \hline
  &     5 & Constant  & 0.0640 & 0.0040 & 15.8701 & 0.0000 \\ 
  &     5 & \textit{rel.type}=Creation & -0.0036 & 0.0011 & -3.3020 & 0.0010 \\ 
  &     5 & \textit{rel.type}=Destruction & 0.0297 & 0.0016 & 18.6011 & 0.0000 \\ 
  &     5 & \textit{neg}=TRUE & -0.0026 & 0.0023 & -1.1385 & 0.2549 \\ 
  &     5 & \textit{fct}=Other & -0.0405 & 0.0040 & -10.0968 & 0.0000 \\ 
  &     5 & \textit{fct}=Remainer & -0.0416 & 0.0049 & -8.5345 & 0.0000 \\ \hline
  &     6 & Constant  & 0.0193 & 0.0020 & 9.5565 & 0.0000 \\ 
  &     6 & \textit{rel.type}=Creation & 0.0033 & 0.0006 & 5.1408 & 0.0000 \\ 
  &     6 & \textit{rel.type}=Destruction & 0.0042 & 0.0009 & 4.6849 & 0.0000 \\ 
  &     6 & \textit{neg}=TRUE & -0.0039 & 0.0016 & -2.4214 & 0.0155 \\ 
  &     6 & \textit{fct}=Other & -0.0028 & 0.0020 & -1.4046 & 0.1601 \\ 
  &     6 & \textit{fct}=Remainer & -0.0034 & 0.0031 & -1.1154 & 0.2647 \\ \hline
  &     7 & Constant  & 0.0106 & 0.0025 & 4.2538 & 0.0000 \\ 
  &     7 & \textit{rel.type}=Creation & -0.0005 & 0.0009 & -0.6007 & 0.5480 \\ 
  &     7 & \textit{rel.type}=Destruction & 0.0421 & 0.0014 & 29.5135 & 0.0000 \\ 
  &     7 & \textit{neg}=TRUE & -0.0081 & 0.0022 & -3.6805 & 0.0002 \\ 
  &     7 & \textit{fct}=Other & 0.0085 & 0.0025 & 3.4044 & 0.0007 \\ 
  &     7 & \textit{fct}=Remainer & 0.0057 & 0.0037 & 1.5426 & 0.1229 \\ \hline
  &     8 & Constant  & 0.0246 & 0.0029 & 8.4942 & 0.0000 \\ 
  &     8 & \textit{rel.type}=Creation & 0.0046 & 0.0009 & 5.2425 & 0.0000 \\ 
  &     8 & \textit{rel.type}=Destruction & -0.0051 & 0.0010 & -4.9672 & 0.0000 \\ 
  &     8 & \textit{neg}=TRUE & 0.0008 & 0.0021 & 0.3784 & 0.7051 \\ 
  &     8 & \textit{fct}=Other & -0.0059 & 0.0029 & -2.0451 & 0.0408 \\ 
  &     8 & \textit{fct}=Remainer & -0.0146 & 0.0036 & -4.0617 & 0.0000 \\ \hline
  &     9 & Constant  & 0.0388 & 0.0027 & 14.2883 & 0.0000 \\ 
  &     9 & \textit{rel.type}=Creation & -0.0175 & 0.0009 & -18.9675 & 0.0000 \\ 
  &     9 & \textit{rel.type}=Destruction & -0.0034 & 0.0012 & -2.8020 & 0.0051 \\ 
  &     9 & \textit{neg}=TRUE & -0.0043 & 0.0023 & -1.8385 & 0.0660 \\ 
  &     9 & \textit{fct}=Other & 0.0063 & 0.0028 & 2.2785 & 0.0227 \\ 
  &     9 & \textit{fct}=Remainer & 0.0091 & 0.0040 & 2.2853 & 0.0223 \\ \hline
  &    10 & Constant  & 0.0252 & 0.0026 & 9.6413 & 0.0000 \\ 
  &    10 & \textit{rel.type}=Creation & 0.0080 & 0.0007 & 10.8194 & 0.0000 \\ 
  &    10 & \textit{rel.type}=Destruction & 0.0006 & 0.0010 & 0.5435 & 0.5868 \\ 
  &    10 & \textit{neg}=TRUE & 0.0018 & 0.0023 & 0.7796 & 0.4356 \\ 
  &    10 & \textit{fct}=Other & -0.0057 & 0.0026 & -2.2104 & 0.0271 \\ 
  &    10 & \textit{fct}=Remainer & -0.0150 & 0.0034 & -4.4076 & 0.0000 \\ \hline
  &    11 & Constant  & 0.0398 & 0.0036 & 11.1730 & 0.0000 \\ 
  &    11 & \textit{rel.type}=Creation & -0.0275 & 0.0013 & -20.4315 & 0.0000 \\ 
  &    11 & \textit{rel.type}=Destruction & -0.0334 & 0.0016 & -20.6136 & 0.0000 \\ 
  &    11 & \textit{neg}=TRUE & 0.0139 & 0.0039 & 3.5857 & 0.0003 \\ 
  &    11 & \textit{fct}=Other & 0.0149 & 0.0035 & 4.2055 & 0.0000 \\ 
  &    11 & \textit{fct}=Remainer & 0.0237 & 0.0051 & 4.6690 & 0.0000 \\ \hline
   &    12 & Constant  & 0.0228 & 0.0023 & 9.9259 & 0.0000 \\ 
  &    12 & \textit{rel.type}=Creation & -0.0010 & 0.0008 & -1.3111 & 0.1898 \\ 
   &    12 & \textit{rel.type}=Destruction & -0.0071 & 0.0011 & -6.2938 & 0.0000 \\ 
   &    12 & \textit{neg}=TRUE & 0.0121 & 0.0027 & 4.5758 & 0.0000 \\ 
   &    12 & \textit{fct}=Other & -0.0018 & 0.0023 & -0.7822 & 0.4341 \\ 
   &    12 & \textit{fct}=Remainer & 0.0016 & 0.0036 & 0.4491 & 0.6534 \\ \hline
   &    13 & Constant  & 0.0132 & 0.0020 & 6.6517 & 0.0000 \\ 
   &    13 & \textit{rel.type}=Creation & 0.0144 & 0.0007 & 21.3358 & 0.0000 \\ 
   &    13 & \textit{rel.type}=Destruction & 0.0021 & 0.0008 & 2.6756 & 0.0075 \\ 
   &    13 & \textit{neg}=TRUE & -0.0044 & 0.0018 & -2.4675 & 0.0136 \\ 
   &    13 & \textit{fct}=Other & -0.0031 & 0.0019 & -1.6165 & 0.1060 \\ 
   &    13 & \textit{fct}=Remainer & 0.0213 & 0.0030 & 7.0397 & 0.0000 \\ \hline
   &    14 & Constant  & 0.0148 & 0.0021 & 7.0230 & 0.0000 \\ 
   &    14 & \textit{rel.type}=Creation & 0.0073 & 0.0007 & 10.6451 & 0.0000 \\ 
   &    14 & \textit{rel.type}=Destruction & -0.0007 & 0.0008 & -0.8895 & 0.3738 \\ 
   &    14 & \textit{neg}=TRUE & -0.0042 & 0.0016 & -2.6802 & 0.0074 \\ 
   &    14 & \textit{fct}=Other & -0.0030 & 0.0020 & -1.5128 & 0.1303 \\ 
   &    14 & \textit{fct}=Remainer & -0.0004 & 0.0031 & -0.1306 & 0.8961 \\ \hline
   &    15 & Constant  & 0.0393 & 0.0025 & 15.5097 & 0.0000 \\ 
   &    15 & \textit{rel.type}=Creation & -0.0235 & 0.0009 & -26.9684 & 0.0000 \\ 
   &    15 & \textit{rel.type}=Destruction & -0.0225 & 0.0012 & -18.8189 & 0.0000 \\ 
   &    15 & \textit{neg}=TRUE & 0.0122 & 0.0024 & 5.1409 & 0.0000 \\ 
  &    15 & \textit{fct}=Other & 0.0001 & 0.0025 & 0.0584 & 0.9534 \\ 
   &    15 & \textit{fct}=Remainer & -0.0067 & 0.0035 & -1.9169 & 0.0553 \\ \hline
   &    16 & Constant  & 0.0547 & 0.0031 & 17.4128 & 0.0000 \\ 
   &    16 & \textit{rel.type}=Creation & 0.0092 & 0.0008 & 11.6683 & 0.0000 \\ 
   &    16 & \textit{rel.type}=Destruction & 0.0081 & 0.0012 & 6.9380 & 0.0000 \\ 
   &    16 & \textit{neg}=TRUE & -0.0026 & 0.0022 & -1.1575 & 0.2471 \\ 
   &    16 & \textit{fct}=Other & -0.0267 & 0.0030 & -8.8965 & 0.0000 \\ 
   &    16 & \textit{fct}=Remainer & -0.0357 & 0.0039 & -9.1678 & 0.0000 \\ \hline
   &    17 & Constant  & 0.0287 & 0.0026 & 11.2354 & 0.0000 \\ 
   &    17 & \textit{rel.type}=Creation & 0.0081 & 0.0008 & 10.7616 & 0.0000 \\ 
   &    17 & \textit{rel.type}=Destruction & 0.0178 & 0.0011 & 16.1294 & 0.0000 \\ 
   &    17 & \textit{neg}=TRUE & -0.0061 & 0.0018 & -3.4714 & 0.0005 \\ 
   &    17 & \textit{fct}=Other & -0.0114 & 0.0026 & -4.3874 & 0.0000 \\ 
   &    17 & \textit{fct}=Remainer & -0.0095 & 0.0033 & -2.8660 & 0.0042 \\ \hline
   &    18 & Constant  & 0.0396 & 0.0035 & 11.4103 & 0.0000 \\ 
   &    18 & \textit{rel.type}=Creation & -0.0032 & 0.0011 & -2.9508 & 0.0032 \\ 
   &    18 & \textit{rel.type}=Destruction & 0.0003 & 0.0014 & 0.2240 & 0.8228 \\ 
   &    18 & \textit{neg}=TRUE & 0.0101 & 0.0027 & 3.7413 & 0.0002 \\ 
   &    18 & \textit{fct}=Other & -0.0083 & 0.0035 & -2.3435 & 0.0191 \\ 
   &    18 & \textit{fct}=Remainer & -0.0111 & 0.0043 & -2.5963 & 0.0094 \\ \hline
   &    19 & Constant  & 0.0285 & 0.0034 & 8.3680 & 0.0000 \\ 
   &    19 & \textit{rel.type}=Creation & -0.0088 & 0.0011 & -7.9786 & 0.0000 \\ 
   &    19 & \textit{rel.type}=Destruction & -0.0102 & 0.0014 & -7.2765 & 0.0000 \\ 
   &    19 & \textit{neg}=TRUE & -0.0085 & 0.0026 & -3.3223 & 0.0009 \\ 
   &    19 & \textit{fct}=Other & 0.0035 & 0.0033 & 1.0435 & 0.2967 \\ 
   &    19 & \textit{fct}=Remainer & 0.0108 & 0.0043 & 2.5015 & 0.0124 \\ \hline
   &    20 & Constant  & 0.0412 & 0.0032 & 12.9566 & 0.0000 \\ 
   &    20 & \textit{rel.type}=Creation & 0.0220 & 0.0010 & 22.1303 & 0.0000 \\ 
   &    20 & \textit{rel.type}=Destruction & 0.0026 & 0.0012 & 2.1172 & 0.0342 \\ 
   &    20 & \textit{neg}=TRUE & 0.0233 & 0.0033 & 7.0589 & 0.0000 \\ 
   &    20 & \textit{fct}=Other & -0.0167 & 0.0032 & -5.1865 & 0.0000 \\ 
   &    20 & \textit{fct}=Remainer & -0.0160 & 0.0038 & -4.2169 & 0.0000 \\ \hline
   &    21 & Constant  & 0.0171 & 0.0026 & 6.4564 & 0.0000 \\ 
   &    21 & \textit{rel.type}=Creation & 0.0207 & 0.0010 & 21.0634 & 0.0000 \\ 
   &    21 & \textit{rel.type}=Destruction & 0.0092 & 0.0012 & 7.8525 & 0.0000 \\ 
  &    21 & \textit{neg}=TRUE & -0.0117 & 0.0023 & -5.1587 & 0.0000 \\ 
  &    21 & \textit{fct}=Other & 0.0002 & 0.0026 & 0.0704 & 0.9438 \\ 
  &    21 & \textit{fct}=Remainer & 0.0006 & 0.0037 & 0.1568 & 0.8754 \\ \hline
  &    22 & Constant  & 0.0288 & 0.0029 & 9.9498 & 0.0000 \\ 
  &    22 & \textit{rel.type}=Creation & -0.0247 & 0.0011 & -23.0179 & 0.0000 \\ 
  &    22 & \textit{rel.type}=Destruction & -0.0200 & 0.0014 & -14.4377 & 0.0000 \\ 
  &    22 & \textit{neg}=TRUE & -0.0050 & 0.0024 & -2.0774 & 0.0378 \\ 
  &    22 & \textit{fct}=Other & 0.0071 & 0.0029 & 2.4084 & 0.0160 \\ 
  &    22 & \textit{fct}=Remainer & -0.0061 & 0.0038 & -1.6008 & 0.1094 \\ \hline
  &    23 & Constant  & 0.0238 & 0.0034 & 7.0785 & 0.0000 \\ 
  &    23 & \textit{rel.type}=Creation & 0.0298 & 0.0012 & 25.9190 & 0.0000 \\ 
  &    23 & \textit{rel.type}=Destruction & 0.0000 & 0.0014 & 0.0041 & 0.9967 \\ 
   &    23 & \textit{neg}=TRUE & -0.0119 & 0.0028 & -4.3161 & 0.0000 \\ 
   &    23 & \textit{fct}=Other & 0.0022 & 0.0034 & 0.6454 & 0.5186 \\ 
   &    23 & \textit{fct}=Remainer & 0.0017 & 0.0046 & 0.3641 & 0.7158 \\ \hline
   &    24 & Constant  & 0.0354 & 0.0029 & 12.3901 & 0.0000 \\ 
   &    24 & \textit{rel.type}=Creation & 0.0157 & 0.0008 & 20.1294 & 0.0000 \\ 
   &    24 & \textit{rel.type}=Destruction & 0.0047 & 0.0011 & 4.3537 & 0.0000 \\ 
  &    24 & \textit{neg}=TRUE & 0.0195 & 0.0026 & 7.4187 & 0.0000 \\ 
  &    24 & \textit{fct}=Other & -0.0065 & 0.0028 & -2.3026 & 0.0213 \\ 
   &    24 & \textit{fct}=Remainer & -0.0074 & 0.0035 & -2.1327 & 0.0330 \\ \hline
   &    25 & Constant  & 0.0221 & 0.0023 & 9.5598 & 0.0000 \\ 
   &    25 & \textit{rel.type}=Creation & -0.0010 & 0.0007 & -1.4067 & 0.1595 \\ 
   &    25 & \textit{rel.type}=Destruction & 0.0142 & 0.0012 & 12.1276 & 0.0000 \\ 
   &    25 & \textit{neg}=TRUE & -0.0070 & 0.0019 & -3.6501 & 0.0003 \\ 
   &    25 & \textit{fct}=Other & 0.0030 & 0.0023 & 1.3099 & 0.1902 \\ 
   &    25 & \textit{fct}=Remainer & 0.0059 & 0.0031 & 1.9232 & 0.0545 \\ \hline
   &    26 & Constant  & 0.0395 & 0.0032 & 12.1924 & 0.0000 \\ 
   &    26 & \textit{rel.type}=Creation & -0.0303 & 0.0012 & -24.6338 & 0.0000 \\ 
   &    26 & \textit{rel.type}=Destruction & -0.0192 & 0.0017 & -11.1510 & 0.0000 \\ 
   &    26 & \textit{neg}=TRUE & -0.0037 & 0.0030 & -1.2225 & 0.2215 \\ 
   &    26 & \textit{fct}=Other & 0.0131 & 0.0032 & 4.1555 & 0.0000 \\ 
   &    26 & \textit{fct}=Remainer & 0.0066 & 0.0043 & 1.5310 & 0.1258 \\ \hline
   &    27 & Constant  & 0.0171 & 0.0024 & 7.0375 & 0.0000 \\ 
   &    27 & \textit{rel.type}=Creation & -0.0028 & 0.0009 & -3.2232 & 0.0013 \\ 
   &    27 & \textit{rel.type}=Destruction & -0.0076 & 0.0011 & -6.6079 & 0.0000 \\ 
   &    27 & \textit{neg}=TRUE & -0.0040 & 0.0021 & -1.9116 & 0.0559 \\ 
   &    27 & \textit{fct}=Other & 0.0073 & 0.0024 & 3.0478 & 0.0023 \\ 
   &    27 & \textit{fct}=Remainer & 0.0066 & 0.0034 & 1.9263 & 0.0541 \\ \hline
   &    28 & Constant  & 0.0284 & 0.0033 & 8.5978 & 0.0000 \\ 
   &    28 & \textit{rel.type}=Creation & 0.0228 & 0.0011 & 21.6190 & 0.0000 \\ 
   &    28 & \textit{rel.type}=Destruction & -0.0040 & 0.0013 & -3.1698 & 0.0015 \\ 
   &    28 & \textit{neg}=TRUE & 0.0187 & 0.0036 & 5.1774 & 0.0000 \\ 
   &    28 & \textit{fct}=Other & -0.0030 & 0.0034 & -0.8910 & 0.3729 \\ 
   &    28 & \textit{fct}=Remainer & 0.0014 & 0.0043 & 0.3351 & 0.7376 \\ \hline
   &    29 & Constant  & 0.0211 & 0.0023 & 9.3169 & 0.0000 \\ 
   &    29 & \textit{rel.type}=Creation & 0.0030 & 0.0006 & 5.1266 & 0.0000 \\ 
   &    29 & \textit{rel.type}=Destruction & -0.0030 & 0.0008 & -3.7888 & 0.0002 \\ 
   &    29 & \textit{neg}=TRUE & -0.0013 & 0.0017 & -0.7728 & 0.4397 \\ 
   &    29 & \textit{fct}=Other & -0.0087 & 0.0022 & -3.8932 & 0.0001 \\ 
   &    29 & \textit{fct}=Remainer & -0.0016 & 0.0029 & -0.5623 & 0.5739 \\ \hline
   &    30 & Constant  & 0.0132 & 0.0019 & 6.9358 & 0.0000 \\ 
   &    30 & \textit{rel.type}=Creation & 0.0031 & 0.0007 & 4.1969 & 0.0000 \\ 
   &    30 & \textit{rel.type}=Destruction & -0.0005 & 0.0010 & -0.5347 & 0.5929 \\ 
   &    30 & \textit{neg}=TRUE & -0.0034 & 0.0018 & -1.8495 & 0.0644 \\ 
   &    30 & \textit{fct}=Other & 0.0064 & 0.0018 & 3.4516 & 0.0006 \\ 
   &    30 & \textit{fct}=Remainer & -0.0041 & 0.0025 & -1.5969 & 0.1103 \\ \hline
   &    31 & Constant  & 0.0367 & 0.0029 & 12.5020 & 0.0000 \\ 
   &    31 & \textit{rel.type}=Creation & -0.0219 & 0.0011 & -20.8501 & 0.0000 \\ 
   &    31 & \textit{rel.type}=Destruction & -0.0253 & 0.0013 & -19.0093 & 0.0000 \\ 
   &    31 & \textit{neg}=TRUE & -0.0013 & 0.0030 & -0.4191 & 0.6752 \\ 
   &    31 & \textit{fct}=Other & 0.0063 & 0.0029 & 2.1692 & 0.0301 \\ 
   &    31 & \textit{fct}=Remainer & -0.0018 & 0.0041 & -0.4389 & 0.6607 \\ \hline
   &    32 & Constant  & 0.0315 & 0.0026 & 12.2313 & 0.0000 \\ 
   &    32 & \textit{rel.type}=Creation & -0.0247 & 0.0009 & -26.9047 & 0.0000 \\ 
   &    32 & \textit{rel.type}=Destruction & 0.0189 & 0.0014 & 13.8506 & 0.0000 \\ 
   &    32 & \textit{neg}=TRUE & -0.0053 & 0.0025 & -2.1541 & 0.0312 \\ 
   &    32 & \textit{fct}=Other & 0.0113 & 0.0026 & 4.4257 & 0.0000 \\ 
   &    32 & \textit{fct}=Remainer & 0.0278 & 0.0046 & 6.0781 & 0.0000 \\ \hline
   &    33 & Constant  & 0.0075 & 0.0025 & 2.9559 & 0.0031 \\ 
   &    33 & \textit{rel.type}=Creation & 0.0230 & 0.0009 & 26.6673 & 0.0000 \\ 
   &    33 & \textit{rel.type}=Destruction & 0.0296 & 0.0013 & 23.6445 & 0.0000 \\ 
   &    33 & \textit{neg}=TRUE & -0.0117 & 0.0021 & -5.4961 & 0.0000 \\ 
   &    33 & \textit{fct}=Other & 0.0108 & 0.0025 & 4.3929 & 0.0000 \\ 
   &    33 & \textit{fct}=Remainer & 0.0213 & 0.0038 & 5.6308 & 0.0000 \\ \hline
   &    34 & Constant  & 0.0232 & 0.0022 & 10.5312 & 0.0000 \\ 
   &    34 & \textit{rel.type}=Creation & 0.0100 & 0.0007 & 15.1427 & 0.0000 \\ 
   &    34 & \textit{rel.type}=Destruction & 0.0009 & 0.0010 & 0.9221 & 0.3565 \\ 
   &    34 & \textit{neg}=TRUE & 0.0021 & 0.0021 & 1.0075 & 0.3137 \\ 
   &    34 & \textit{fct}=Other & -0.0034 & 0.0022 & -1.5807 & 0.1140 \\ 
   &    34 & \textit{fct}=Remainer & -0.0021 & 0.0029 & -0.7178 & 0.4729 \\ \hline
   &    35 & Constant  & 0.0109 & 0.0021 & 5.2107 & 0.0000 \\ 
   &    35 & \textit{rel.type}=Creation & 0.0189 & 0.0008 & 24.1147 & 0.0000 \\ 
   &    35 & \textit{rel.type}=Destruction & 0.0021 & 0.0009 & 2.1888 & 0.0286 \\ 
   &    35 & \textit{neg}=TRUE & -0.0056 & 0.0020 & -2.7687 & 0.0056 \\ 
   &    35 & \textit{fct}=Other & 0.0074 & 0.0021 & 3.4534 & 0.0006 \\ 
   &    35 & \textit{fct}=Remainer & 0.0009 & 0.0031 & 0.2990 & 0.7649 \\ \hline
   &    36 & Constant  & 0.0069 & 0.0021 & 3.2615 & 0.0011 \\ 
   &    36 & \textit{rel.type}=Creation & 0.0127 & 0.0008 & 15.9608 & 0.0000 \\ 
   &    36 & \textit{rel.type}=Destruction & 0.0062 & 0.0011 & 5.8633 & 0.0000 \\ 
   &    36 & \textit{neg}=TRUE & -0.0073 & 0.0019 & -3.8505 & 0.0001 \\ 
   &    36 & \textit{fct}=Other & 0.0098 & 0.0021 & 4.7018 & 0.0000 \\ 
   &    36 & \textit{fct}=Remainer & 0.0179 & 0.0033 & 5.4723 & 0.0000 \\ \hline
   &    37 & Constant  & 0.0302 & 0.0023 & 13.1362 & 0.0000 \\ 
   &    37 & \textit{rel.type}=Creation & -0.0262 & 0.0009 & -30.1670 & 0.0000 \\ 
   &    37 & \textit{rel.type}=Destruction & -0.0225 & 0.0012 & -18.9386 & 0.0000 \\ 
   &    37 & \textit{neg}=TRUE & -0.0001 & 0.0023 & -0.0411 & 0.9672 \\ 
   &    37 & \textit{fct}=Other & 0.0118 & 0.0023 & 5.0894 & 0.0000 \\ 
   &    37 & \textit{fct}=Remainer & 0.0131 & 0.0035 & 3.7157 & 0.0002 \\ \hline
   &    38 & Constant  & 0.0074 & 0.0004 & 16.5572 & 0.0000 \\ 
   &    38 & \textit{rel.type}=Creation & 0.0002 & 0.0002 & 1.0797 & 0.2803 \\ 
   &    38 & \textit{rel.type}=Destruction & -0.0003 & 0.0002 & -1.2309 & 0.2184 \\ 
   &    38 & \textit{neg}=TRUE & -0.0001 & 0.0004 & -0.3539 & 0.7234 \\ 
   &    38 & \textit{fct}=Other & 0.0039 & 0.0005 & 8.6132 & 0.0000 \\ 
   &    38 & \textit{fct}=Remainer & 0.0016 & 0.0006 & 2.5763 & 0.0100 \\ \hline
\end{longtable}
\FloatBarrier

\clearpage
\section{Robustness Checks}\label{S_sec:Robustness}

\begin{table}[]
    \centering
        \caption{Parameters used in experiments with different values for K from 3 to 70}
    \label{S_tab:K_choice_parameters}
    \begin{tabular}{|c|c|}
    \hline
    \textbf{parameter} & \textbf{value}\\\hline
       K  & c(3:70)\\
       Number of experiment replications  &  50 \\
       Topic prevalence &  ~ $rel.type + neg + fct + s(t)$\\
       Topic content & ~ $fct$\\
        Maximum Expectation Maximization Iterations &  200\\
        Expectations Maximization Tolerance & 1e-05\\
        Initialization Type & Spectral\\
        Proportion of docs to be held out & 25\%\\
        M & 10\\\hline
    \end{tabular}
\vspace{2em}
\end{table}

\begin{figure}
\includegraphics[trim= 0mm 20mm 0mm 15mm,clip, width=0.9\textwidth]{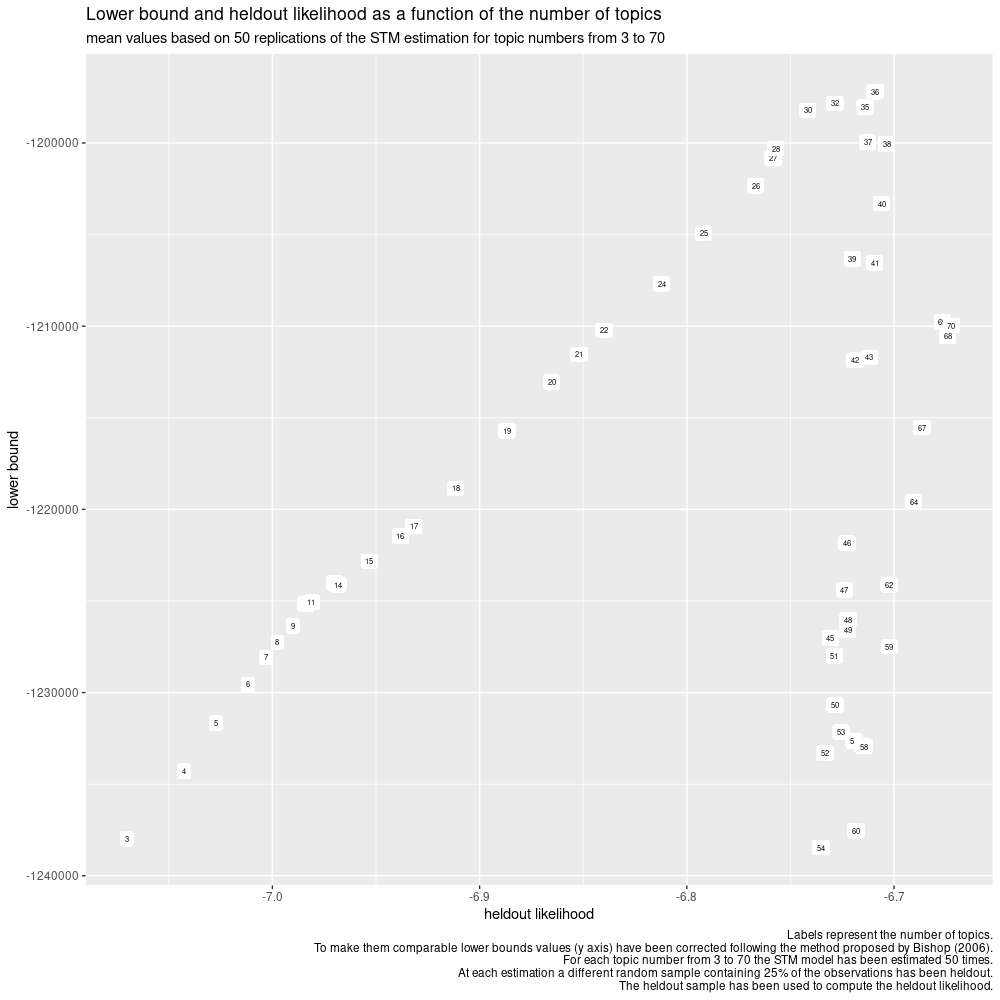}
\caption{Mean values of  \textit{Lower bound}  and \textit{Heldout likelihood} (based on 50 replications) of estimated STM  different values of $K$. Labels represent the number of topics $K$.\newline
At each replication a different random sample containing a 25\%  of the population of no-deal effects has been heldout. The heldout sample was hence used to compute the \textit{Heldout likelihood} .}
\label{S_fig:heldout_lb_likelihood}
\end{figure}

\clearpage

\section{Methodology}\label{S_sec:Methodology}

\begin{table}
\caption{List of stopwords  (based on Quanteda stopword list) used in STM estimation step}
    \label{S_tab:stopwords}
    \centering
\begin{tabular}{|c|}
    \hline
    \textbf{stopwords}     \\\hline
 "i", "me", "my", "myself","we","our","ours",  "ourselves", "you", "your","yours","yourself", "yourselves","he",\\
 "him","his","himself","she","her","hers","herself","it","its","itself","they","them","their", "theirs" ,"themselves",\\
 "what","which","who" ,"whom",  "this" ,"that" ,"these", "those", "am", "is" ,"are","was","were", "be","been",\\
 "being","have" ,"has" ,"had","having","do","does","did","doing","would","should","could","ought","i'm",\\
 "you're","he's","she's","it's","we're","they're","i've","you've","we've", "they've","i'd",   "you'd", "he'd",  "she'd",\\
 "we'd",  "they'd",  "i'll",  "you'll",  "he'll", "she'll",  "we'll", "they'll", "isn't", "aren't",  "wasn't",  "weren't",\\ 
 "hasn't",  "haven't", "hadn't",  "doesn't", "don't", "didn't",  "won't", "wouldn't"   "shan't",  "shouldn't"  "can't",\\
\hline
\end{tabular}
\end{table}

\begin{table}
\caption{RegEx used to identify causes with "no-deal" in the last four tokens of the cause-side}
    \label{S_tab:nodeal_regex}
    \centering
    \begin{tabular}{|c|}
    \hline
    \textbf{RegEx}     \\\hline
{\scriptsize ([Nn]\{1\}[Oo]\{1\}[-\_[:space:]]\{1\}[Dd]\{1\}[Ee]\{1\}[Aa]\{1\}[lL]\{1\})\{1\}[\textbackslash{}"'?\textgreater{}]\{0,1\}([[:space:]]+[[A-Za-z\_]]+)\{0,3\}[[:space:][:punct:]]\{0,\}\textdollar{}} \\\hline
\end{tabular}
\end{table}

\begin{table}
  \caption{ \label{S_tab:verbslist}List of causal verbs employed in RegEx functions by relation verb type ($rel.type$)}
    \tiny
  \begin{tabular}{|c|c|c|}

  \hline
 \textit{Destruction }&  \textit{Causation} &  \textit{Creation}\\\hline
destroy/destroys/destroyed/destroying   &  cause/causes/caused/causing  &   create/creates/creating/created \\
demolish/demolishes/demolished/demolishing  &  determine/determines/determined/determining &  produce/produces/produced/producing  \\
annihilate/annihilates/annihilated/annihilating  &  engender/engenders/engendered/engendering &  originate/originates/originated/originating \\
obliterate/obliterates/obliterated/obliterating  &   entail/entails/entailed/entailing &   generate/generates/generated/generating \\
cancel/cancels/cancelled/canceled/cancelling/canceling   &  provoke/provokes/provoked/provoking & make/makes/made/making  \\
eradicate/eradicates/eradicated/eradicating        &   trigger/triggers/triggered/triggering & \\
extirpate/extirpates/extirpated/extirpating & leads to/lead to/led to/leading to &    \\
  eliminate/eliminates/eliminated/eliminating    & result in/results in/resulted in/resulting in & \\
wipe out/wipes out/wiped out/wiping out &  & \\
 wipe off/wipes off/wiped off/wiping off &  & \\ 
kill/kills/killed/killing &  & \\ \hline
  \end{tabular}
\end{table}

%what about "reason of" 
%  "due to","led by","result from","results from", "resulted from","resulting from", "stem from", "stems from", "stemmed from", "stemming from", "derive from", "derives from", "derived from", "deriving from", "originate from", "originates from", "originated from", "originating from"
\pagebreak
\newpage
\begin{table}
\vspace{1em}
    \caption{\label{S_tab:rel_extraction_RegEx}%
Relation extraction function and associated regular expressions (R code)
}
\vspace{-2em}
\end{table}
%%

\bibliographystyle{chicago}  % plain apalike siam chicago ecta
\bibliography{bibliography.bib}